\documentclass[sigconf,nonacm]{acmart}

\usepackage{booktabs}
\usepackage{multirow}
\usepackage{makecell}
\usepackage{graphicx}
\usepackage{subcaption}
\usepackage{mathtools}

\usepackage{microtype}
\usepackage{xurl}
\usepackage[most]{tcolorbox}
\tcbuselibrary{listings,breakable}
\usepackage[ruled,vlined,linesnumbered]{algorithm2e}

\allowdisplaybreaks
\setcellgapes{2pt}
\makegapedcells

\renewcommand\footnotetextcopyrightpermission[1]{}

\begin{document}

\title{Uncertainty-Guided LLM Semantic Augmentation for Heterogeneous Treatment Effect Estimation}

\author{Jialu Xu}
\authornote{Jialu Xu and Mengkun Liang contributed equally to this work.}
\affiliation{%
  \institution{MIIT Key Laboratory of Data and Decision Intelligence, Beihang University}
  \city{Beijing}
  \country{China}
}
\email{xujialu08@buaa.edu.cn}

\author{Mengkun Liang}
\authornotemark[1]
\affiliation{%
  \institution{MIIT Key Laboratory of Data and Decision Intelligence, Beihang University}
  \city{Beijing}
  \country{China}
}
\email{mengkun@buaa.edu.cn}

\author{Guannan Liu}
\authornote{Corresponding author.}
\affiliation{%
  \institution{MIIT Key Laboratory of Data and Decision Intelligence, Beihang University}
  \city{Beijing}
  \country{China}
}
\email{liugn@buaa.edu.cn}

\author{Xiaojie Mao}
\affiliation{%
  \institution{School of Economics and Management, Tsinghua University}
  \city{Beijing}
  \country{China}
}
\email{maoxj@sem.tsinghua.edu.cn}

\author{Junjie Wu}
\affiliation{%
  \institution{MIIT Key Laboratory of Data and Decision Intelligence, Beihang University}
  \city{Beijing}
  \country{China}
}
\email{wujj@buaa.edu.cn}

\begin{abstract}
Estimating heterogeneous treatment effects is central to targeted interventions, such as personalized promotions and precision medicine. We focus on the conditional average treatment effect (CATE), a standard estimand for characterizing such heterogeneity. Even under standard identification conditions, finite-sample CATE estimation requires learning the nuisance structure for covariate adjustment and treatment-effect heterogeneity, often together with an effective representation of $X$. Raw numerical and categorical encodings can leave semantic relations and higher-order interactions implicit, making this joint task locally unstable. A motivating study further shows that this instability appears through partially separable assignment- and heterogeneity-side channels. Building on this observation, we propose CURL (Causal Uncertainty-guided Representation Learning), a plug-in adapter that uses estimator uncertainty to allocate pretrained semantic capacity to locally unstable units. CURL queries a frozen LLM through two role-conditioned prompts, constructs assignment- and heterogeneity-oriented representations from the observed covariates, and routes them through separated pathways. On four benchmarks, CURL improves ten host learners in most settings, while ablation, refinement-dynamics, route-reassignment, and probe analyses support the intended design and roles of the two channels.
\end{abstract}


\maketitle

\section{Introduction}
Estimating heterogeneous treatment effects is central to individualized interventions in precision medicine, targeted marketing, and policy evaluation; we focus on the conditional average treatment effect (CATE), which characterizes how expected effects vary with observed covariates \cite{shalit2017estimating,kunzel2019metalearners,nie2021quasi}. Reliable CATE estimation requires the covariates $X$ to support confounding adjustment and to reveal systematic variation in treatment response \cite{imbens2015causal}.
Estimation can fail for two reasons: relevant individual-level causal information may be absent from $X$, threatening identification, or, despite identification, the raw encoding of
$X$ may make adjustment and heterogeneity structure difficult for finite-sample learners to exploit.

In this paper, we focus on the second and less-examined failure mode. We term the gap between the raw covariate encoding and a task-effective representation for finite-sample CATE learning \emph{representational lossiness}. The notion is relative to the encoding, learner, and sample size, rather than an intrinsic information loss in $X$ or a failure of identification. For example, diagnosis codes may omit semantic proximity, while customer categories may obscure behavioral patterns that jointly shape targeting and response. Meta-learning, balanced-representation, and latent-variable methods provide flexible tools for CATE estimation \cite{kunzel2019metalearners,shalit2017estimating,shi2019adapting,louizos2017causal}, but learn representations and associated functions mainly from the observed sample. When semantic relations and higher-order interactions remain implicit, limited data cannot effectively share statistical strength across related observations, leaving local estimates unstable. This motivates an external inductive bias that reorganizes observed information without adding new individual-level facts.

\begin{figure}[t]
    \centering
    \includegraphics[width=\columnwidth]{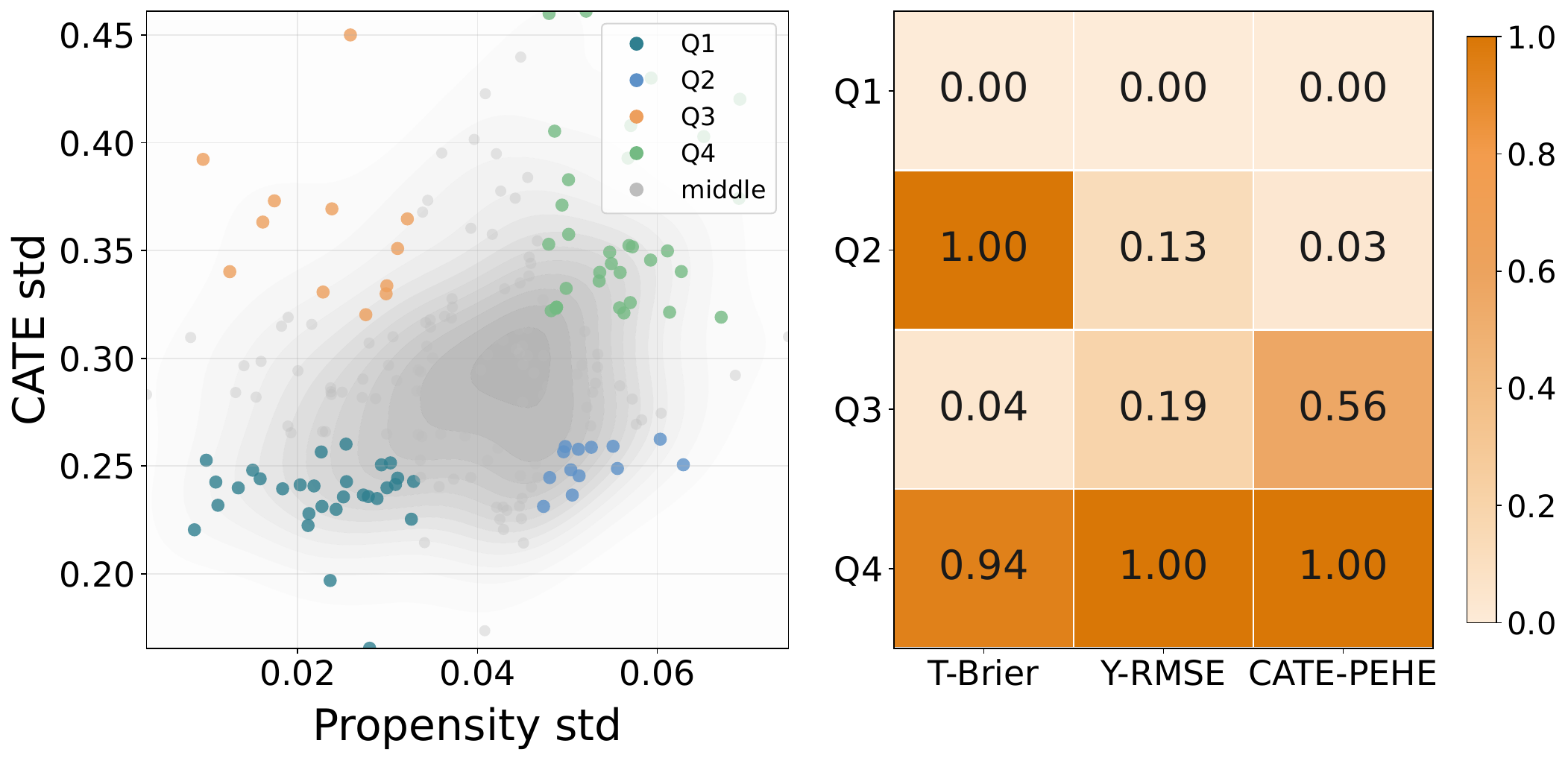}
    \caption{Motivating study on IHDP using an S-Learner with an auxiliary propensity head. The left panel places samples on the propensity--CATE uncertainty plane. Q1--Q4 are the four corner groups, and the rest form the middle group. The right panel reports column-normalized T-Brier, Y-RMSE, and CATE-PEHE. Q2 is dominated by assignment error, Q3 by effect-estimation error, and Q4 by both.}
    \label{fig:preliminary}
\end{figure}

Large language models (LLMs) offer such an inductive bias because pretraining captures broad semantic regularities that may be difficult to learn from a single finite observational sample, including in structured tabular prediction~\citep{wen2024generative}. Prior work has examined LLMs for causal reasoning and discovery \cite{kiciman2023causal,du2025causal}, and recent estimators have used LLM predictions with textual confounders or unstructured records \cite{veljanovski2024doublelingo,dhawan2024natural,chen2024proximal}. However, how pretrained semantic knowledge should be incorporated into CATE estimators for structured covariates remains open. Applying a common embedding to every sample is costly and can introduce irrelevant variation, while an undifferentiated representation may mix information needed for assignment modeling with information needed for heterogeneous response. A useful design must therefore determine where semantic capacity is needed, construct representations for distinct statistical roles, and route them into the appropriate components of the estimator.

Figure~\ref{fig:preliminary} provides empirical support for a selective and role-conditioned design. We fit an S-Learner with an auxiliary propensity head on IHDP, derive CATE from its treatment-conditioned outcome predictions, and estimate both uncertainties via Monte Carlo (MC) dropout \cite{gal2016dropout}. Q2 has the largest normalized treatment Brier error but low CATE error, whereas Q3 shows the opposite pattern; Q4 is difficult on both dimensions. These results indicate that local unreliability varies across samples and cannot be captured by a single score. Propensity uncertainty reflects instability in treatment-selection modeling relevant to covariate adjustment, whereas CATE uncertainty reflects instability in treatment-response heterogeneity. We therefore use both as operational allocation signals rather than evidence of missing causal variables, consistent with prior work on uncertainty-based identification of unreliable treatment-effect estimates \cite{jesson2020identifying}.

Motivated by these empirical patterns, we propose CURL, an uncertainty-guided representation adapter for CATE estimation under \textit{representational lossiness}. CURL uses propensity and CATE uncertainty to prioritize a fixed proportion of locally unstable samples for LLM-based semantic augmentation, then constructs independent assignment-oriented and heterogeneity-oriented representations from their observed profiles. The former refines the shared covariate representation, while the latter enters the effect component as a separate feature block and is excluded from propensity estimation. The augmented set is updated as the host estimator evolves. In this way, CURL expands the effective hypothesis class without changing the observed information set, target estimand, or identifying assumptions.

Our contributions are summarized as follows.
\begin{itemize}
    \item We distinguish representational lossiness from causal information insufficiency, characterizing it as a finite-sample bottleneck in exploiting semantic relations left implicit by raw covariate encodings.
    \item We develop CURL, a plug-in adapter that uses uncertainty to allocate assignment- and heterogeneity-oriented semantic representations and integrates them through asymmetric routing, progressive refinement, and training-calibrated test-time inference.
    \item We evaluate CURL across four benchmarks and ten CATE estimators, with analyses of predictive performance, robustness, refinement dynamics, and semantic-channel roles.
\end{itemize}

\section{Related Work}
\label{sec:related_work}

\subsection{Representation Learning for CATE Estimation}

CATE estimation has developed along three major paradigms. \emph{Meta-learning} methods reduce effect estimation to supervised prediction, including the S-, T-, and X-learner~\citep{kunzel2019metalearners} and the R-learner~\citep{nie2021quasi}. \emph{Balanced-representation} methods learn treatment-invariant representations, as in TARNet, CFRNet~\citep{shalit2017estimating}, and DragonNet~\citep{shi2019adapting}. \emph{Deep latent-variable} methods infer hidden factors from proxies in $X$, including CEVAE~\citep{louizos2017causal}, TEDVAE~\citep{Zhang2021tedvae}, $\beta$-Intact-VAE~\citep{wu2022beta}, CFDiVAE~\citep{xu2024causal}, and iVAE~\citep{pmlr-v108-khemakhem20a}, with further extensions to structure uncertainty, adjustment-feature selection, and disentangled or robust representations~\citep{tran2022structure,wang2023adjustment,zhong2022descn,li2024self,wang2024cercfr}. Despite these advances, most representations are learned primarily from the study sample and may remain unreliable when the encoding of $X$ is poorly aligned with the estimator's effective hypothesis class. Sensitivity to model misspecification and proxy construction further motivates using pretrained semantic structure to help finite-sample estimators exploit information already contained in $X$~\citep{rissanen2021critical,zhu2025causal}.

\subsection{LLMs for Causal Augmentation}

Recent work has explored LLMs for causal analysis along two lines. On the \emph{qualitative-structure} side, LLMs have been used for causal discovery and reasoning from text~\citep{kiciman2023causal,liu2025large}, causal graph
construction~\citep{ban2025llm,du2025causal}, and formal causal-query
benchmarks~\citep{NEURIPS2023_631bb943}, although current models may still rely on shallow Rung-1 associations~\citep{chi2024unveiling}. On the \emph{quantitative-estimation} side, DoubleLingo uses LLM-based nuisance models with text confounders~\citep{veljanovski2024doublelingo}; NATURAL leverages LLM-derived conditional information from text~\citep{dhawan2024natural}; proximal text-based inference extracts proxies for proximal identification~\citep{chen2024proximal}; and GATE generates counterfactual outcomes for small-sample structured data~\citep{huynh2025improving}. These studies primarily use LLMs to process textual inputs, construct auxiliary variables, or directly predict causal quantities. Less attention has been paid to using pretrained semantic structure as an inductive bias that helps CATE estimators better exploit structured covariates under limited data.

\subsection{Uncertainty Estimation in Causal Inference}

Epistemic uncertainty has long been studied in Bayesian machine learning, from the Laplace approximation~\citep{mackay1992practical} to MC dropout~\citep{gal2016dropout}, deep ensembles~\citep{lakshminarayanan2017simple},
and variational inference~\citep{blundell2015weight}, with
\citet{kendall2017uncertainties} distinguishing aleatoric and epistemic
uncertainty. In decision-making, uncertainty supports exploration through UCB-style bandits and Thompson sampling~\citep{abbasi2011improved,zhou2020neural,xu2021neural},
or conservatism in offline settings~\citep{an2021uncertainty,bai2022pessimistic,wu2021uncertainty}.
Closer to causal estimation, \citet{zhang2023uncertainty} account for logging-policy uncertainty in MSE-optimal importance weighting, whereas CATE methods mainly use uncertainty for confidence quantification, failure detection, or budgeted sample acquisition~\citep{alaa2017bayesian,jesson2020identifying,wen2025progressive}. In contrast, CURL uses uncertainty to allocate semantic augmentation to locally unstable samples.

\section{Problem Formulation}
\label{sec:problem}

Let $\mathcal D_n=\{(X_i,T_i,Y_i)\}_{i=1}^n\sim P^n$ be an i.i.d. observational sample, where $X_i\in\mathcal X\subseteq\mathbb R^d$ contains pretreatment covariates, $T_i\in\{0,1\}$ is a binary treatment, and $Y_i=T_iY_i(1)+(1-T_i)Y_i(0)$. We study the CATE, \textit{i.e.}, $\tau(x)=\mathbb E[Y(1)-Y(0)\mid X=x]$.

We assume conditional unconfoundedness $\{Y(0),Y(1)\}\perp T\mid X$ and overlap $\epsilon\leq e(x)\leq1-\epsilon$, where $e(x)=\Pr(T=1\mid X=x)$~\citep{imbens2015causal}. Because treatment assignment may depend on $X$ in observational data, $e(x)$ summarizes systematic treatment-selection differences across the covariate space and, depending on the estimator, supports weighting, balancing, or residualization for covariate adjustment~\citep{rosenbaum1983central,nie2021quasi}. Together with consistency, these assumptions identify $\tau(x)=\mu_1(x)-\mu_0(x)$, where $\mu_t(x)=\mathbb E[Y\mid T=t,X=x]$. Thus, $X$ is sufficient for causal adjustment, and unmeasured confounding is outside our scope.

Let $\mathcal H$ be the function class of a host estimator. Its best available CATE approximation under the raw encoding of $X$ is
\begin{equation}
\tau_{\mathcal H}^{*}
\in
\arg\min_{f\in\mathcal H}
\mathbb E_{X\sim P_X}
\left[
\left\{
f(X)-\tau(X)
\right\}^{2}
\right].
\label{eq:hte_objective}
\end{equation}

Identification alone does not ensure that $\tau_{\mathcal H}^{*}$ is easy to express or learn from finite data. A learner must estimate the assignment and outcome structure needed for adjustment and how treatment effects vary with $X$, often while learning an effective representation from the same sample. When the raw numerical and categorical encoding of $X$ leaves semantic relations and higher-order interactions implicit, these functions may become unnecessarily complex within the host class, limiting statistical sharing across related observations and destabilizing local estimates. This is precisely the \emph{representational lossiness} introduced in Section 1. Concretely, although $X$ identifies $\tau(x)$, relevant adjustment and heterogeneity structure remains difficult to recover from its raw encoding. The gap is thus relative to the encoding, learner, and sample size, rather than an intrinsic information loss in $X$ or a failure of identification.

\section{Methodology}
\label{sec:method}

\subsection{Framework Overview}
\label{sec:method-overview}

To address the \textit{representational lossiness}, \textbf{CURL} (\underline{C}ausal \underline{U}ncertainty-guided \underline{R}epresentation \underline{L}earning) augments a host estimator $F_\theta$ with a plug-in representation adapter that makes adjustment- and heterogeneity-related structure in $X$ easier to exploit under finite data. CURL preserves the host-specific prediction structure and main objective, while introducing adapter parameters $\eta$ for the semantic projectors, gates, routing components, and any auxiliary propensity head. We write $\Theta=(\theta,\eta)$ for the joint trainable state and $F_\Theta$ for the wrapped estimator. Host-specific instantiations are provided in Appendix~\ref{appendix:adapters}.

\begin{figure*}[t]
    \centering
    \includegraphics[width=0.8\textwidth]{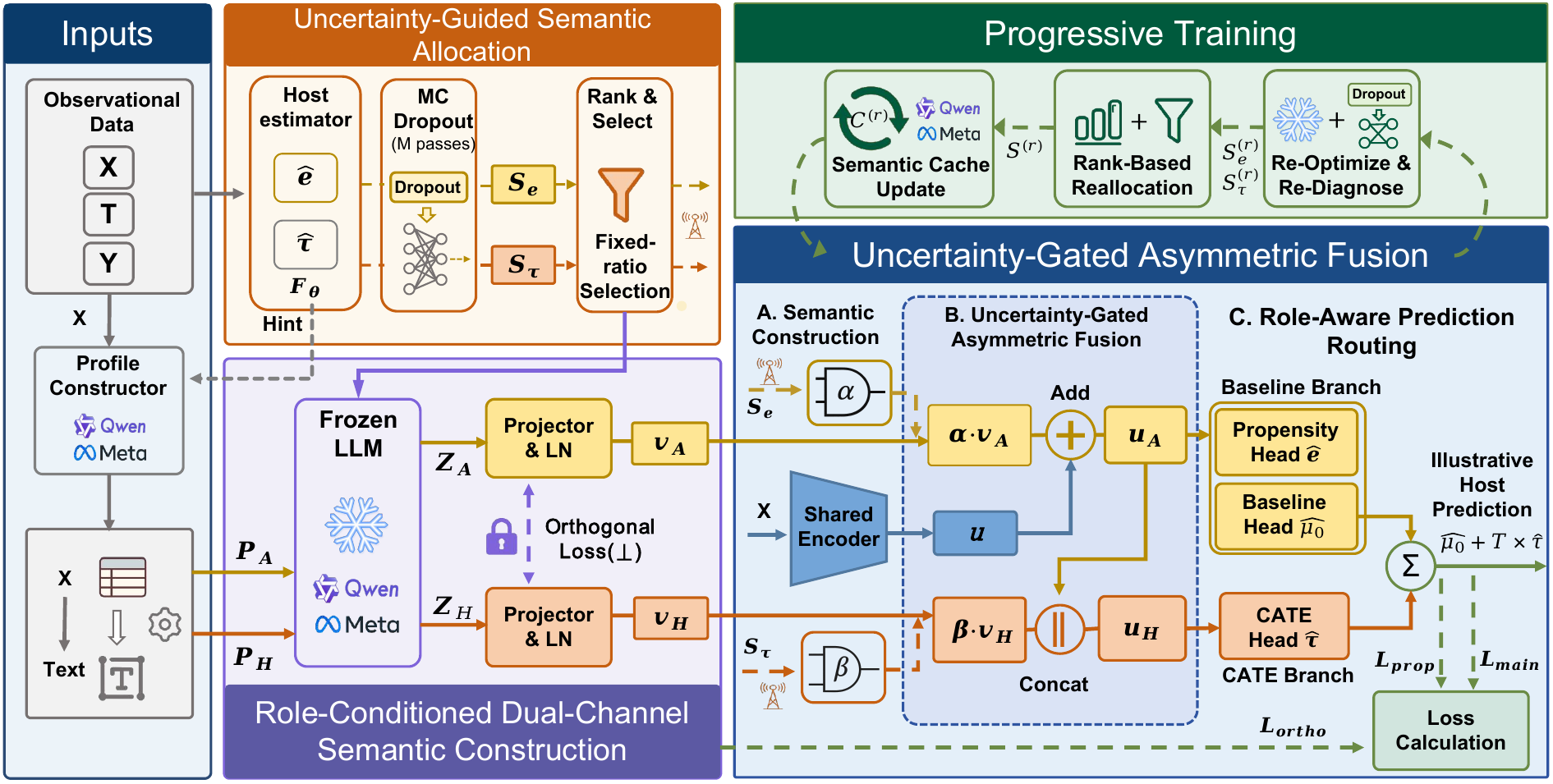}
    \caption{Overall architecture of CURL. Uncertainty-guided diagnosis
    identifies samples that receive semantic augmentation;
    dual-channel LLM encoding constructs assignment- and
    heterogeneity-oriented representations; and role-aware integration
    routes them into the corresponding components of the host
    estimator. The procedure progressively refreshes the uncertainty
    diagnosis as the wrapped estimator evolves.}
    \label{fig:architecture}
\end{figure*}

As illustrated in Figure~\ref{fig:architecture}, CURL operates through four coupled mechanisms. \emph{Uncertainty-Guided Semantic Allocation} (Section~\ref{sec:method-diagnostic}) uses two epistemic-uncertainty scores, one for the propensity component and one for the CATE component, to prioritize units with high local predictive instability. \emph{Role-Conditioned Semantic Augmentation and Routing} (Section~\ref{sec:method-disentangle}) queries a frozen LLM independently for each selected unit to build an assignment-oriented and a heterogeneity-oriented representation from its observed covariates. \emph{Role-Aware Prediction Routing} (Section~\ref{sec:method-adapt}) injects the two representations into the host through separated pathways, so that heterogeneity-oriented information never reaches the propensity component. \emph{Progressive Refinement and Test-Time Inference} (Section~\ref{sec:method-training}) repeats this diagnosis as $\Theta$ evolves, and a calibrated procedure extends the same allocation and routing rules to unseen units at test time.

We index training rounds by $r\in\{0,\ldots,R-1\}$ and write $\Theta^{(r)}$ and $F^{(r)}:=F_{\Theta^{(r)}}$ for the parameter state and wrapped estimator at the start of round $r$; this superscript is used consistently in what follows. Because $\eta$ only augments, and never replaces, the host's own computation, $F^{(0)}$ with no cached semantic input reduces exactly to the unaugmented host $F_\theta$. Next, we describe the four modules in detail.

\subsection{Uncertainty-Guided Semantic Allocation}
\label{sec:method-diagnostic}

To diagnose the current estimator $F^{(r)}$, we keep its normalization layers in evaluation mode and dropout layers active, and perform $M$ stochastic forward passes using MC dropout~\cite{gal2016dropout}. Writing $\widehat e_i^{(r,m)}$ and $\widehat\tau_i^{(r,m)}$ for the propensity and CATE predictions obtained in the $m$-th pass for unit $i$, we take the standard deviation of each prediction across the $M$ passes as the corresponding uncertainty score,
\begin{equation}
s_{e,i}^{(r)}
=
\operatorname{Std}_{m=1}^{M}\!\left[\widehat e_i^{(r,m)}\right],
\qquad
s_{\tau,i}^{(r)}
=
\operatorname{Std}_{m=1}^{M}\!\left[\widehat\tau_i^{(r,m)}\right].
\label{eq:sig_defs}
\end{equation}
These two scores capture local predictive instability separately for the assignment component and the treatment effect component.

At the initial allocation round, some host estimators do not provide a native propensity output. To apply the same allocation rule across all hosts and rounds, we set $s_{e,i}^{(0)}=0$ for all units in this case. After wrapping, CURL introduces and trains a propensity head for every host. Consequently, from the first refinement round onward, both $s_{e,i}^{(r)}$ and $s_{\tau,i}^{(r)}$ are available and jointly determine semantic allocation.

CURL converts the two scores into a single ranking through their empirical percentile ranks. Let $\operatorname{PRank}(a_i)$ denote the percentile rank of $a_i$ among $\{a_j\}_{j=1}^{N}$, and let $\rho\in(0,1]$ be the per-round selection ratio. The units prioritized for semantic allocation in round $r$ are
\begin{equation}
\mathcal S^{(r)}
=
\operatorname{Top}_{\lfloor \rho N\rfloor}
\left\{
\operatorname{PRank}\!\left(s_{e,i}^{(r)}\right)
+
\operatorname{PRank}\!\left(s_{\tau,i}^{(r)}\right)
:
i=1,\ldots,N
\right\},
\label{eq:hard_iter}
\end{equation}
with ties broken deterministically so that the selection budget stays fixed. Taking percentile ranks places the two signals on a common scale despite their different units and magnitudes, while their sum provides a unified ranking over both uncertainty sources. When $s_{e,i}^{(r)}$ is set to zero for all units under the exception above, the ranking reduces to CATE-side uncertainty alone.

Because selections may overlap across rounds, previously queried units reuse their cached semantic embeddings. Let $\mathcal S^{(r)}$ denote the units selected in round $r$, $\mathcal C^{(r)}$ those cached by the end of that round, and $\mathcal Q^{(r)}$ the selected but uncached units. With $\mathcal C^{(-1)}=\varnothing$,

\begin{equation}
\mathcal C^{(r)}
=
\mathcal C^{(r-1)}\cup\mathcal S^{(r)},
\qquad
\mathcal Q^{(r)}
=
\mathcal S^{(r)}\setminus\mathcal C^{(r-1)}.
\label{eq:cache_update}
\end{equation}

Thus, the LLM is queried only for $\mathcal Q^{(r)}$, so each training unit is queried at most once.

\subsection{Role-Conditioned Semantic Augmentation and Routing}
\label{sec:method-disentangle}
After determining which units require additional semantic capacity in Section~\ref{sec:method-diagnostic}, the next question is \emph{what} semantic information these units should receive and \emph{where} it should act within the estimator. To answer this, we introduce the notion of a \emph{role}, \textit{i.e.}, the function a semantic channel serves in CATE estimation: assignment modeling or treatment-effect heterogeneity.
Accordingly, this section specifies what role-conditioned information the selected units should receive, how strongly it is integrated, and which prediction pathways it may influence.

\subsubsection{Role-Conditioned Dual-Channel Semantic Construction}
The units in $\mathcal Q^{(r)}$ identified in the previous step are the ones for which CURL queries the LLM. Each such unit is first rendered into a natural-language profile $p_i=\psi(x_i)$ by a deterministic, dataset-specific constructor $\psi$. A dataset-level hint $h_{\mathrm{ins}}$ is generated once from aggregate covariate differences between the initially selected and full training populations, and is then held fixed for the remainder of training. Neither $p_i$ nor $h_{\mathrm{ins}}$ contains unit-level treatment, outcome, counterfactual, ground-truth effect, or test statistics; full field mappings, aggregation rules, and templates are deferred to Appendix~\ref{app:prompt-instances}.

From this profile, two role-specific constructors produce an assignment-oriented prompt $P_{A,i}=P_A(p_i,h_{\mathrm{ins}})$, eliciting pre-treatment context for assignment and outcomes, and a heterogeneity-oriented prompt $P_{H,i}=P_H(p_i,h_{\mathrm{ins}})$, eliciting stable attributes for treatment sensitivity. These functional roles target distinct estimator pathways rather than alternative descriptions of the unit. The prompts use independent LLM calls and caches, preventing either from becoming an autoregressive continuation of the other. For either prompt $P\in\{P_{A,i},P_{H,i}\}$ described in Appendix~\ref{app:prompt-template}, CURL extracts the final-layer contextual hidden state at the last valid response token, before the output head. The resulting channel embeddings are
\begin{equation}
z_{A,i}^{\mathrm{LLM}}
=
\operatorname{Emb}_{\Phi}(P_{A,i}),
\qquad
z_{H,i}^{\mathrm{LLM}}
=
\operatorname{Emb}_{\Phi}(P_{H,i}),
\label{eq:emb}
\end{equation}
and are cached permanently once computed. For either channel $q\in\{A,H\}$, the sparse input supplied to the adapter in round $r$ zeroes out the cached embedding for any unit not yet in $\mathcal C^{(r)}$,
\begin{equation}
z_{q,i}^{(r)}
=
\mathbb I\!\left\{i\in\mathcal C^{(r)}\right\}
z_{q,i}^{\mathrm{LLM}}.
\label{eq:sparse_mem}
\end{equation}
Units outside $\mathcal C^{(r)}$ therefore receive zero semantic input and follow the unaugmented host pathway.

\subsubsection{Uncertainty-Gated Asymmetric Fusion}
The host encoder produces a base representation for each unit, $u_i^{(r)}=f_{\theta^{(r)}}(x_i)\in\mathbb R^{d_h}$, using only the host parameters $\theta^{(r)}$ and evolving across rounds as $\theta^{(r)}$ is updated. Two trainable projectors $g_A$ and $g_H$ adapt the sparse semantic inputs into $d_h$-dimensional features used by the wrapped estimator,
\begin{equation}
v_{A,i}^{(r)}=g_A(z_{A,i}^{(r)}),
\qquad
v_{H,i}^{(r)}=g_H(z_{H,i}^{(r)}).
\label{eq:projections}
\end{equation}
The strength with which each channel is injected is controlled by a scalar gate. Each gate combines the matching uncertainty score, passed through a small network and the sigmoid function $\sigma(a)=(1+e^{-a})^{-1}$, with an indicator restricting injection to units that have actually been queried,
\begin{equation}
\begin{aligned}
\alpha_i^{(r)}
&=
\mathbb I\!\left\{i\in\mathcal C^{(r)}\right\}
\sigma\!\left(
\operatorname{MLP}_A(s_{e,i}^{(r)})
\right),\\
\beta_i^{(r)}
&=
\mathbb I\!\left\{i\in\mathcal C^{(r)}\right\}
\sigma\!\left(
\operatorname{MLP}_H(s_{\tau,i}^{(r)})
\right).
\end{aligned}
\label{eq:gates}
\end{equation}
The cache indicator blocks unqueried units, while each gate aligns a role with its matching uncertainty: $s_e$ modulates assignment semantics and $s_\tau$ modulates heterogeneity semantics. Writing $\operatorname{Concat}(a,b)$ for feature-wise concatenation, CURL forms assignment-adjusted shared and effect-sensitive representations as

\begin{align}
u_{A,i}^{(r)}&=u_i^{(r)}+\alpha_i^{(r)}v_{A,i}^{(r)},
\label{eq:r_clean}\\
u_{H,i}^{(r)}&=\operatorname{Concat}\!\left(u_{A,i}^{(r)},\beta_i^{(r)}v_{H,i}^{(r)}\right).
\label{eq:h_fused}
\end{align}
The fusion rules serve distinct roles. Residual addition treats assignment semantics as a correction to the shared state, preserving its dimension and recovering the host representation when inactive. Concatenation retains heterogeneity semantics as a separate block that the CATE operator can parameterize independently. This asymmetry is a role-specific routing bias, not a claim that the embeddings form an identifiable causal decomposition; their redundancy is further discouraged by the orthogonality penalty in Section~\ref{sec:method-training}.

\subsubsection{Role-Aware Prediction Routing}
\label{sec:method-adapt}
The propensity head $g_{e,\Theta^{(r)}}$ of the current wrapped estimator consumes only the assignment-adjusted shared representation, while its host-specific CATE operator $\mathcal T_{\Theta^{(r)}}$ consumes $u_{H,i}^{(r)}$, which already contains $u_{A,i}^{(r)}$ through the concatenation in Eq.~\eqref{eq:h_fused},
\begin{equation}
\widehat e_{\Theta^{(r)}}(x_i)
=
g_{e,\Theta^{(r)}}\!\left(u_{A,i}^{(r)}\right), \quad
\widehat\tau_{\Theta^{(r)}}(x_i)
=
\mathcal T_{\Theta^{(r)}}\!\left(u_{H,i}^{(r)}\right).
\label{eq:role_routing}
\end{equation}
The two roles are thus enforced structurally: heterogeneity semantics cannot reach assignment prediction, whereas the CATE operator accesses both the assignment-adjusted shared state and effect-specific representation. This interface covers treatment-conditioned single-head, two-potential-outcome-head, direct-CATE, and latent-variable hosts; their insertion points and native objectives appear in Appendix~\ref{appendix:adapters}.

\subsection{Progressive Refinement and Test-Time Inference}
\label{sec:method-training}

\subsubsection{Progressive Training}
Training begins by fitting the unaugmented host with its native objective, which we write as $\operatorname{FitHost}(\mathcal D)$, giving $F^{(0)}=\operatorname{FitHost}(\mathcal D)$ and an empty cache $\mathcal C^{(-1)}=\varnothing$. Each subsequent round $r=0,\ldots,R-1$ repeats the diagnosis and allocation steps of Eqs.~\eqref{eq:sig_defs}--\eqref{eq:cache_update}, queries the frozen LLM only on the newly surfaced set $\mathcal Q^{(r)}$, and resumes optimization from the current parameters to obtain $F^{(r+1)}$, without reinitializing either the estimator or its optimizer. Within such an optimization block, the cached embeddings and the frozen LLM $\Phi$ are held fixed, while the host, the projectors, the gates, and the prediction heads are trained jointly; the uncertainty scores produced at the end of a block then serve as the gate inputs for the next one, so that training advances through the same cycle of diagnosis, allocation, and integration introduced in Section~\ref{sec:method-overview}.

All trainable parameters within a block are updated by minimizing a single objective that adds two auxiliary terms to the host-side prediction loss $\mathcal L_{\mathrm{main}}$, which already includes any host-specific balancing, latent-variable, or reconstruction term. Writing $\operatorname{BCE}(p,t)$ for binary cross-entropy and $\cos(a,b)=\langle a,b\rangle/(\lVert a\rVert_2\lVert b\rVert_2+\epsilon)$ for cosine similarity with numerical stabilizer $\epsilon>0$, the two auxiliary terms supervise the propensity head and discourage redundancy between the two semantic channels,
\begin{align}
\mathcal L(\Theta)
&=\mathcal L_{\mathrm{main}}
+\lambda_{\mathrm{prop}}\mathcal L_{\mathrm{prop}}
+\lambda_{\mathrm{ortho}}\mathcal L_{\mathrm{ortho}},
\label{eq:overall_loss}\\
\mathcal L_{\mathrm{prop}}
&=\frac{1}{N}\sum_{i=1}^{N}
\operatorname{BCE}\!\left(\widehat e_{\Theta^{(r)}}(x_i),t_i\right),
\label{eq:l_prop}\\
\mathcal L_{\mathrm{ortho}}
&=\frac{1}{|\mathcal C^{(r)}|}
\sum_{i\in\mathcal C^{(r)}}
\cos^2\!\left(v_{A,i}^{(r)},v_{H,i}^{(r)}\right),
\label{eq:l_ortho}
\end{align}
with $\lambda_{\mathrm{prop}},\lambda_{\mathrm{ortho}}\geq0$ and $\mathcal L_{\mathrm{ortho}}=0$ when $\mathcal C^{(r)}$ is empty. The auxiliary propensity head is present in every evaluated CURL adapter, so setting $\lambda_{\mathrm{prop}}=0$ recovers the same objective form for a host that lacks this interface natively. Algorithmic details are given in Appendix~\ref{app:algorithm}.

\subsubsection{Test-time Inference}
Test-time inference follows a calibration-and-refinement principle. CURL converts the relative ranking used during training into a fixed pointwise rule for deciding whether an unseen unit should receive semantic augmentation, then recomputes uncertainty after augmentation to control semantic fusion. This conversion is necessary because Eq.~\eqref{eq:hard_iter} relies on relative rankings over the full training set, whereas test-time decisions must be made pointwise without assuming access to other test units or depending on the composition of an arbitrary test batch.

Let $s_{e,i}^{(R)}$ and $s_{\tau,i}^{(R)}$ denote the two uncertainty scores computed at the final training state, following the same definition as Eq.~\eqref{eq:sig_defs}. Their empirical distribution functions,
\begin{equation}
\widehat G_e^{\mathrm{tr}}(s)=\frac{1}{N}\sum_{i=1}^{N}\mathbb I(s_{e,i}^{(R)}\le s),
\qquad
\widehat G_\tau^{\mathrm{tr}}(s)=\frac{1}{N}\sum_{i=1}^{N}\mathbb I(s_{\tau,i}^{(R)}\le s),
\label{eq:train_cdf}
\end{equation}
map the two scores, which may have different scales, to percentile positions within the final training uncertainty distributions. Summing these percentiles gives $\omega_i^{\mathrm{tr}}=\widehat G_e^{\mathrm{tr}}(s_{e,i}^{(R)})+\widehat G_\tau^{\mathrm{tr}}(s_{\tau,i}^{(R)})$, and CURL stores their empirical $(1-\rho)$-quantile, $\kappa_\rho=Q_{1-\rho}({\omega_i^{\mathrm{tr}}}_{i=1}^{N})$, together with the two distributions above. Using the same $\rho$ calibrates this threshold to the training-time selection budget and gives a comparable augmentation rate when the test uncertainty distribution is similar.

For an unseen covariate vector $x_*$, CURL first sets both semantic inputs to zero and computes pre-augmentation uncertainties $s_{e,*}^{\mathrm{pre}}$ and $s_{\tau,*}^{\mathrm{pre}}$ through the same Monte Carlo procedure as Eq.~\eqref{eq:sig_defs}. It queries the LLM if and only if
\begin{equation}
a_*=\mathbb I\!\left(
\widehat G_e^{\mathrm{tr}}(s_{e,*}^{\mathrm{pre}})
+\widehat G_\tau^{\mathrm{tr}}(s_{\tau,*}^{\mathrm{pre}})
\ge\kappa_\rho\right),
\label{eq:test_decision}
\end{equation}
where $a_*\in\{0,1\}$ is the augmentation decision. A selected unit receives the two embeddings constructed in Section~\ref{sec:method-disentangle}, whereas an unselected unit retains zero embeddings. CURL then recomputes uncertainty under the updated inputs and uses the post-augmentation scores as the gate inputs of Eq.~\eqref{eq:gates} before applying Eq.~\eqref{eq:role_routing}. Thus, the pre-augmentation scores determine whether the unit is sufficiently unstable relative to the final trained estimator to justify an LLM query, while the post-augmentation scores determine how strongly the resulting embeddings influence prediction. No test-time treatment, outcome, ground-truth effect, or comparison across other test units enters either step.

\section{Experiments}
\label{sec:experiments}

We evaluate CURL on four benchmarks and ten CATE estimators, asking: (RQ1) whether its gains are consistent across estimators and robust to backbone and hyperparameter choices; (RQ2) which components drive these gains; and (RQ3) whether the two semantic channels exhibit their intended assignment- and heterogeneity-oriented roles.

\subsection{Experimental Setup}
\label{subsec:experimental_setup}

\paragraph{Datasets and metrics.}
We use two semi-synthetic benchmarks with ground-truth individual effects, IHDP and Adult, evaluated by PEHE, $\epsilon_{\mathrm{ATE}}$, and $\epsilon_{\mathrm{ATT}}$ (lower is better). Jobs and the randomized Hillstrom email experiment lack individual-level effect ground truth and are evaluated by AUUC, AUQC, and LIFT@30 (higher is better). Dataset and metric details are provided in Appendices~\ref{app:exp_details} and~\ref{app:metrics}.

\paragraph{Baselines.}
We plug \textsc{CURL} into ten host learners spanning meta-learners, balanced-representation methods, and deep latent-variable methods (full list in Appendix~\ref{app:exp_details_impl}). We additionally
compare with two LLM-only baselines, \textit{LLM} and
\textit{LLM+MLP}, and with \textit{GATE}~\citep{huynh2025improving},
which directly imputes counterfactual outcomes. Feature-disentangled CATE baselines are reported in Appendix~\ref{app:additional_baseline}.

\paragraph{Implementation.}
We select Qwen2.5-7B or Llama-3-8B per dataset by validation and compare four additional backbones in Section~\ref{subsec:sensitivity_analysis}. We set $\rho=0.25$, $M=30$, $R=3$, $\lambda_{\mathrm{prop}}=1.0$, and $\lambda_{\mathrm{ortho}}=0.1$, and report means over five random seeds. Further details are provided in Appendices~\ref{app:exp_details_impl} and~\ref{app:prompts}.

\newcommand{\ms}[2]{#1\,{\scriptsize$\pm$\,#2}}
\newcommand{\bestms}[2]{\textbf{#1}\,{\scriptsize$\pm$\,\textbf{#2}}}
\newcommand{\gain}[1]{#1\%}

\begin{table*}[t]
\centering
\caption{Results on real-world uplift benchmarks (AUUC, AUQC, and LIFT@30; higher is better). Mean $\pm$ standard deviation over five runs is reported. For each host model and metric, the best result among \textit{base}, GATE, and \textsc{CURL} is boldfaced. $\Delta$ is the relative gain of \textsc{CURL} over the host learner (\%).}
\label{tab:observational_results}
\setlength{\tabcolsep}{2.2pt}
\renewcommand{\arraystretch}{1.08}
\resizebox{\textwidth}{!}{%
\begin{tabular}{llcccccccccccc}
\toprule
\multirow{2}{*}{Dataset} & \multirow{2}{*}{Base Model}
& \multicolumn{4}{c}{AUUC $\uparrow$}
& \multicolumn{4}{c}{AUQC $\uparrow$}
& \multicolumn{4}{c}{LIFT@30 $\uparrow$} \\
\cmidrule(lr){3-6} \cmidrule(lr){7-10} \cmidrule(lr){11-14}
& & \textit{base} & GATE & \textsc{CURL} & $\Delta$
& \textit{base} & GATE & \textsc{CURL} & $\Delta$
& \textit{base} & GATE & \textsc{CURL} & $\Delta$ \\
\midrule
\multirow{12}{*}{Jobs} & LLM & \ms{336.07}{152.90} & -- & -- & -- & \ms{51.52}{15.64} & -- & -- & -- & \ms{3.25}{1.20} & -- & -- & -- \\
 & LLM+MLP & \ms{810.99}{130.44} & -- & -- & -- & \ms{69.11}{22.21} & -- & -- & -- & \ms{2.56}{1.23} & -- & -- & -- \\
\addlinespace[1pt]
 & S-Learner & \ms{506.40}{210.85} & \ms{461.15}{47.89} & \bestms{522.99}{233.41} & \gain{+3.28} & \ms{262.09}{89.67} & \ms{272.56}{28.26} & \bestms{352.08}{64.15} & \gain{+34.34} & \ms{11.33}{6.78} & \ms{2.88}{0.30} & \bestms{11.53}{4.21} & \gain{+1.76} \\
 & T-Learner & \ms{671.81}{254.99} & \ms{522.02}{157.29} & \bestms{875.93}{299.94} & \gain{+30.38} & \ms{351.35}{113.76} & \ms{397.51}{86.95} & \bestms{487.50}{142.68} & \gain{+38.75} & \ms{6.13}{1.15} & \ms{7.27}{1.58} & \bestms{9.22}{3.93} & \gain{+50.41} \\
 & R-Learner & \ms{770.80}{332.56} & \ms{630.93}{258.18} & \bestms{1007.40}{344.72} & \gain{+30.70} & \ms{226.62}{44.71} & \ms{168.09}{46.37} & \bestms{240.12}{88.37} & \gain{+5.96} & \ms{6.60}{2.71} & \ms{7.68}{0.17} & \bestms{8.14}{3.01} & \gain{+23.32} \\
 & TARNet & \ms{491.85}{70.27} & \ms{489.47}{170.58} & \bestms{885.49}{223.55} & \gain{+80.03} & \ms{218.54}{45.24} & \ms{203.95}{29.86} & \bestms{266.82}{57.53} & \gain{+22.09} & \ms{8.06}{2.80} & \ms{8.47}{1.25} & \bestms{11.73}{2.20} & \gain{+45.65} \\
 & DragonNet & \ms{918.04}{318.45} & \ms{655.31}{204.07} & \bestms{1185.06}{389.74} & \gain{+29.09} & \ms{413.35}{122.60} & \ms{376.23}{140.30} & \bestms{454.69}{64.98} & \gain{+10.00} & \ms{11.51}{3.31} & \ms{6.17}{2.27} & \bestms{12.24}{8.31} & \gain{+6.36} \\
 & CFRNet & \ms{816.48}{79.80} & \ms{521.71}{197.32} & \bestms{819.61}{123.81} & \gain{+0.38} & \ms{223.08}{63.98} & \ms{237.80}{89.07} & \bestms{291.50}{58.45} & \gain{+30.67} & \ms{7.32}{1.17} & \ms{7.67}{1.89} & \bestms{8.65}{1.12} & \gain{+18.18} \\
 & CEVAE & \ms{970.01}{197.59} & \ms{442.97}{125.07} & \bestms{975.08}{180.33} & \gain{+0.52} & \ms{465.46}{56.80} & \ms{472.25}{52.95} & \bestms{578.90}{185.02} & \gain{+24.37} & \ms{9.10}{1.79} & \ms{5.06}{0.33} & \bestms{10.63}{1.85} & \gain{+16.80} \\
 & TEDVAE & \ms{862.90}{436.75} & \ms{696.57}{150.40} & \bestms{1087.30}{253.83} & \gain{+26.01} & \ms{417.32}{95.35} & \ms{372.78}{7.02} & \bestms{420.50}{81.05} & \gain{+0.76} & \ms{4.61}{1.23} & \ms{4.71}{0.16} & \bestms{6.16}{2.82} & \gain{+33.48} \\
 & SDD & \ms{574.98}{180.84} & \ms{710.59}{208.96} & \bestms{743.96}{165.72} & \gain{+29.39} & \ms{217.93}{63.15} & \ms{208.53}{60.82} & \bestms{218.64}{75.71} & \gain{+0.32} & \ms{7.85}{3.16} & \ms{8.99}{2.07} & \bestms{9.47}{3.26} & \gain{+20.70} \\
 & CiVAE & \ms{717.18}{131.83} & \ms{655.36}{177.13} & \bestms{1023.04}{278.45} & \gain{+42.65} & \ms{236.21}{21.96} & \ms{179.78}{112.48} & \bestms{238.78}{66.90} & \gain{+1.09} & \ms{4.10}{2.19} & \ms{4.85}{0.57} & \bestms{5.13}{1.77} & \gain{+25.10} \\
\midrule
\multirow{12}{*}{Hillstrom} & LLM & \ms{0.016}{0.009} & -- & -- & -- & \ms{0.011}{0.001} & -- & -- & -- & \ms{0.858}{0.153} & -- & -- & -- \\
 & LLM+MLP & \ms{0.053}{0.032} & -- & -- & -- & \ms{0.014}{0.006} & -- & -- & -- & \ms{1.345}{0.526} & -- & -- & -- \\
\addlinespace[1pt]
 & S-Learner & \ms{0.112}{0.044} & \ms{0.115}{0.013} & \bestms{0.167}{0.053} & \gain{+48.35} & \ms{0.044}{0.011} & \ms{0.023}{0.004} & \bestms{0.045}{0.012} & \gain{+2.49} & \ms{2.570}{0.405} & \ms{2.853}{2.965} & \bestms{3.239}{0.977} & \gain{+26.04} \\
 & T-Learner & \ms{0.106}{0.035} & \ms{0.131}{0.016} & \bestms{0.136}{0.038} & \gain{+28.01} & \ms{0.020}{0.002} & \ms{0.021}{0.008} & \bestms{0.023}{0.003} & \gain{+12.25} & \ms{1.005}{0.221} & \ms{0.973}{0.157} & \bestms{1.109}{0.239} & \gain{+10.38} \\
 & R-Learner & \ms{0.205}{0.049} & \ms{0.132}{0.052} & \bestms{0.207}{0.049} & \gain{+1.07} & \ms{0.046}{0.003} & \ms{0.044}{0.042} & \bestms{0.049}{0.005} & \gain{+6.75} & \ms{4.273}{0.963} & \ms{2.841}{0.815} & \bestms{5.195}{1.563} & \gain{+21.57} \\
 & TARNet & \ms{0.063}{0.020} & \ms{0.062}{0.016} & \bestms{0.065}{0.022} & \gain{+2.22} & \ms{0.022}{0.002} & \ms{0.020}{0.005} & \bestms{0.026}{0.005} & \gain{+14.67} & \ms{0.991}{0.192} & \ms{0.858}{0.100} & \bestms{1.014}{0.186} & \gain{+2.36} \\
 & DragonNet & \ms{0.109}{0.021} & \ms{0.122}{0.058} & \bestms{0.124}{0.024} & \gain{+13.98} & \ms{0.021}{0.001} & \ms{0.024}{0.008} & \bestms{0.026}{0.004} & \gain{+21.40} & \ms{0.995}{0.163} & \ms{0.865}{0.394} & \bestms{1.099}{0.203} & \gain{+10.49} \\
 & CFRNet & \ms{0.130}{0.041} & \ms{0.107}{0.012} & \bestms{0.162}{0.049} & \gain{+24.27} & \ms{0.037}{0.005} & \ms{0.032}{0.023} & \bestms{0.044}{0.007} & \gain{+19.07} & \ms{2.674}{0.538} & \ms{2.497}{2.546} & \bestms{3.734}{1.399} & \gain{+39.61} \\
 & CEVAE & \ms{0.063}{0.021} & \ms{0.053}{0.016} & \bestms{0.068}{0.015} & \gain{+6.97} & \ms{0.019}{0.002} & \ms{0.014}{0.002} & \bestms{0.021}{0.002} & \gain{+11.46} & \ms{0.973}{0.151} & \ms{0.864}{0.371} & \bestms{1.168}{0.189} & \gain{+20.03} \\
 & TEDVAE & \ms{0.111}{0.035} & \ms{0.113}{0.016} & \bestms{0.116}{0.037} & \gain{+4.23} & \ms{0.021}{0.001} & \ms{0.021}{0.005} & \bestms{0.022}{0.001} & \gain{+2.83} & \ms{0.689}{0.299} & \ms{0.875}{0.187} & \bestms{1.126}{0.034} & \gain{+63.37} \\
 & SDD & \ms{0.115}{0.018} & \ms{0.120}{0.016} & \bestms{0.128}{0.019} & \gain{+10.49} & \ms{0.021}{0.001} & \bestms{0.029}{0.015} & \ms{0.023}{0.003} & \gain{+9.35} & \ms{1.022}{0.111} & \bestms{1.709}{0.540} & \ms{1.028}{0.116} & \gain{+0.60} \\
 & CiVAE & \ms{0.066}{0.022} & \ms{0.062}{0.014} & \bestms{0.073}{0.027} & \gain{+11.13} & \ms{0.030}{0.009} & \ms{0.020}{0.003} & \bestms{0.033}{0.011} & \gain{+11.04} & \ms{1.425}{0.364} & \ms{1.407}{0.598} & \bestms{1.511}{0.367} & \gain{+6.04} \\
\bottomrule
\end{tabular}%
}
\end{table*}

\begin{table*}[t]
\centering
\caption{Results on semi-synthetic datasets (PEHE, $\epsilon_{\mathrm{ATE}}$, and $\epsilon_{\mathrm{ATT}}$; lower is better). Mean $\pm$ standard deviation over five runs is reported. For each host model and metric, the best result among \textit{base}, GATE, and \textsc{CURL} is boldfaced. $\Delta$ is the relative error reduction of \textsc{CURL} over the host learner (\%).}
\label{tab:semi_synthetic_results}
\setlength{\tabcolsep}{2.2pt}
\renewcommand{\arraystretch}{1.08}
\resizebox{\textwidth}{!}{%
\begin{tabular}{llcccccccccccc}
\toprule
\multirow{2}{*}{Dataset} & \multirow{2}{*}{Base Model}
& \multicolumn{4}{c}{PEHE $\downarrow$}
& \multicolumn{4}{c}{$\epsilon_{\mathrm{ATE}}$ $\downarrow$}
& \multicolumn{4}{c}{$\epsilon_{\mathrm{ATT}}$ $\downarrow$} \\
\cmidrule(lr){3-6} \cmidrule(lr){7-10} \cmidrule(lr){11-14}
& & \textit{base} & GATE & \textsc{CURL} & $\Delta$
& \textit{base} & GATE & \textsc{CURL} & $\Delta$
& \textit{base} & GATE & \textsc{CURL} & $\Delta$ \\
\midrule
\multirow{12}{*}{IHDP} & LLM & \ms{44.117}{26.698} & -- & -- & -- & \ms{54.308}{41.443} & -- & -- & -- & \ms{53.266}{40.708} & -- & -- & -- \\
 & LLM+MLP & \ms{2.487}{0.935} & -- & -- & -- & \ms{1.398}{0.389} & -- & -- & -- & \ms{1.231}{0.420} & -- & -- & -- \\
\addlinespace[1pt]
 & S-Learner & \ms{4.616}{0.838} & \ms{3.499}{1.106} & \bestms{2.926}{0.790} & \gain{+36.62} & \ms{3.050}{0.290} & \ms{2.492}{0.350} & \bestms{1.950}{0.221} & \gain{+36.08} & \ms{2.771}{0.289} & \ms{2.394}{0.391} & \bestms{1.437}{0.045} & \gain{+48.15} \\
 & T-Learner & \ms{1.643}{0.416} & \ms{1.450}{0.547} & \bestms{1.438}{0.354} & \gain{+12.45} & \ms{0.614}{0.159} & \ms{0.595}{0.065} & \bestms{0.511}{0.089} & \gain{+16.88} & \ms{0.406}{0.144} & \ms{0.526}{0.136} & \bestms{0.275}{0.051} & \gain{+32.21} \\
 & R-Learner & \ms{2.629}{0.725} & \ms{2.625}{0.890} & \bestms{2.297}{0.284} & \gain{+12.64} & \ms{1.179}{0.323} & \ms{1.184}{0.066} & \bestms{1.163}{0.270} & \gain{+1.40} & \ms{0.956}{0.297} & \ms{0.986}{0.315} & \bestms{0.936}{0.243} & \gain{+2.04} \\
 & TARNet & \ms{1.712}{0.509} & \ms{1.370}{0.385} & \bestms{1.324}{0.254} & \gain{+22.67} & \ms{0.609}{0.227} & \ms{0.436}{0.121} & \bestms{0.200}{0.047} & \gain{+67.26} & \ms{0.478}{0.200} & \ms{0.664}{0.277} & \bestms{0.343}{0.097} & \gain{+28.30} \\
 & DragonNet & \ms{1.654}{0.473} & \ms{1.531}{0.419} & \bestms{1.342}{0.217} & \gain{+18.83} & \ms{0.513}{0.158} & \ms{0.505}{0.077} & \bestms{0.315}{0.055} & \gain{+38.60} & \ms{0.332}{0.120} & \ms{0.550}{0.195} & \bestms{0.225}{0.072} & \gain{+32.26} \\
 & CFRNet & \ms{1.767}{0.632} & \ms{1.797}{0.611} & \bestms{1.405}{0.342} & \gain{+20.47} & \ms{0.412}{0.085} & \ms{0.459}{0.034} & \bestms{0.179}{0.068} & \gain{+56.64} & \ms{0.389}{0.201} & \ms{0.617}{0.127} & \bestms{0.307}{0.056} & \gain{+21.06} \\
 & CEVAE & \ms{1.879}{0.730} & \ms{1.548}{0.499} & \bestms{1.508}{0.402} & \gain{+19.77} & \ms{0.953}{0.477} & \ms{0.937}{0.185} & \bestms{0.328}{0.152} & \gain{+65.61} & \ms{0.841}{0.429} & \ms{0.910}{0.455} & \bestms{0.258}{0.122} & \gain{+69.32} \\
 & TEDVAE & \ms{2.150}{0.314} & \ms{2.409}{0.705} & \bestms{1.990}{0.592} & \gain{+7.46} & \ms{1.410}{0.547} & \ms{1.955}{0.456} & \bestms{1.303}{1.032} & \gain{+7.54} & \ms{0.531}{0.030} & \ms{1.929}{0.706} & \bestms{0.455}{0.039} & \gain{+14.36} \\
 & SDD & \ms{1.875}{0.642} & \ms{1.737}{0.337} & \bestms{1.571}{0.396} & \gain{+16.20} & \ms{0.500}{0.184} & \ms{0.543}{0.091} & \bestms{0.497}{0.085} & \gain{+0.42} & \ms{0.369}{0.069} & \ms{0.544}{0.424} & \bestms{0.336}{0.045} & \gain{+8.89} \\
 & CiVAE & \ms{3.099}{0.924} & \ms{3.096}{1.001} & \bestms{3.011}{0.846} & \gain{+2.85} & \ms{0.231}{0.024} & \ms{0.412}{0.172} & \bestms{0.226}{0.058} & \gain{+2.38} & \ms{0.690}{0.148} & \ms{0.657}{0.166} & \bestms{0.598}{0.184} & \gain{+13.42} \\
\midrule
\multirow{12}{*}{Adult} & LLM & \ms{57.312}{20.847} & -- & -- & -- & \ms{1.569}{0.702} & -- & -- & -- & \ms{0.708}{0.295} & -- & -- & -- \\
 & LLM+MLP & \ms{0.429}{0.063} & -- & -- & -- & \ms{0.191}{0.065} & -- & -- & -- & \ms{0.170}{0.060} & -- & -- & -- \\
\addlinespace[1pt]
 & S-Learner & \ms{0.260}{0.036} & \ms{0.471}{0.112} & \bestms{0.213}{0.041} & \gain{+18.05} & \ms{0.158}{0.048} & \ms{0.454}{0.115} & \bestms{0.095}{0.050} & \gain{+39.91} & \ms{0.135}{0.061} & \ms{0.454}{0.135} & \bestms{0.113}{0.031} & \gain{+16.06} \\
 & T-Learner & \ms{0.993}{0.207} & \ms{1.067}{0.349} & \bestms{0.894}{0.270} & \gain{+9.97} & \ms{0.354}{0.070} & \ms{0.917}{0.293} & \bestms{0.161}{0.103} & \gain{+54.55} & \ms{0.396}{0.037} & \ms{0.877}{0.280} & \bestms{0.281}{0.075} & \gain{+29.17} \\
 & R-Learner & \ms{0.065}{0.012} & \ms{0.153}{0.017} & \bestms{0.062}{0.021} & \gain{+5.35} & \ms{0.005}{0.004} & \ms{0.101}{0.013} & \bestms{0.004}{0.004} & \gain{+20.75} & \ms{0.007}{0.004} & \ms{0.098}{0.021} & \bestms{0.002}{0.001} & \gain{+72.97} \\
 & TARNet & \ms{0.658}{0.088} & \ms{1.049}{0.314} & \bestms{0.649}{0.125} & \gain{+1.32} & \ms{0.331}{0.069} & \ms{0.889}{0.259} & \bestms{0.175}{0.096} & \gain{+46.93} & \ms{0.355}{0.044} & \ms{0.855}{0.248} & \bestms{0.275}{0.072} & \gain{+22.76} \\
 & DragonNet & \ms{0.607}{0.047} & \ms{1.032}{0.334} & \bestms{0.596}{0.059} & \gain{+1.73} & \ms{0.337}{0.049} & \ms{0.867}{0.283} & \bestms{0.146}{0.088} & \gain{+56.72} & \ms{0.346}{0.026} & \ms{0.828}{0.278} & \bestms{0.240}{0.065} & \gain{+30.62} \\
 & CFRNet & \ms{0.947}{0.270} & \ms{0.986}{0.334} & \bestms{0.941}{0.128} & \gain{+0.67} & \ms{0.772}{0.286} & \ms{0.837}{0.284} & \bestms{0.452}{0.260} & \gain{+41.46} & \ms{0.780}{0.308} & \ms{0.804}{0.277} & \bestms{0.491}{0.238} & \gain{+36.98} \\
 & CEVAE & \ms{0.408}{0.062} & \ms{0.653}{0.286} & \bestms{0.386}{0.145} & \gain{+5.41} & \ms{0.203}{0.101} & \ms{0.537}{0.291} & \bestms{0.080}{0.064} & \gain{+60.60} & \ms{0.177}{0.103} & \ms{0.536}{0.315} & \bestms{0.113}{0.067} & \gain{+36.51} \\
 & TEDVAE & \ms{0.443}{0.192} & \ms{0.621}{0.733} & \bestms{0.436}{0.085} & \gain{+1.63} & \ms{0.212}{0.089} & \ms{0.457}{0.679} & \bestms{0.211}{0.052} & \gain{+0.61} & \ms{0.223}{0.123} & \ms{0.435}{0.641} & \bestms{0.221}{0.083} & \gain{+0.90} \\
 & SDD & \bestms{0.642}{0.119} & \ms{1.034}{0.317} & \ms{0.717}{0.216} & \gain{-11.75} & \ms{0.345}{0.080} & \ms{0.857}{0.269} & \bestms{0.334}{0.089} & \gain{+3.16} & \ms{0.339}{0.095} & \ms{0.813}{0.266} & \bestms{0.329}{0.082} & \gain{+2.98} \\
 & CiVAE & \ms{0.326}{0.260} & \ms{0.302}{0.123} & \bestms{0.114}{0.053} & \gain{+65.06} & \ms{0.195}{0.231} & \ms{0.191}{0.132} & \bestms{0.037}{0.025} & \gain{+81.12} & \ms{0.210}{0.218} & \ms{0.194}{0.137} & \bestms{0.041}{0.021} & \gain{+80.64} \\
\bottomrule
\end{tabular}%
}
\end{table*}

\subsection{Main Results}
\label{subsec:main_results}

Tables~\ref{tab:observational_results} and~\ref{tab:semi_synthetic_results} report the main results on the four benchmarks, respectively. We summarize three observations.

First, \textsc{CURL} improves its host learner across the vast majority of settings, with one degradation on Adult. The improvement appears on both benchmarks with ground truth and benchmarks evaluated by uplift metrics, indicating that it is not specific to a particular evaluation protocol.
These patterns are consistent with \textsc{CURL} helping finite-sample estimators exploit semantic structure in $X$ that is difficult to learn from its raw encoding.

Second, the improvement spans meta-learners, balanced-representation methods, and latent-variable methods, supporting the use of \textsc{CURL} as a plug-in adapter rather than an architecture-specific estimator. The gains for latent-variable baselines further suggest that pretrained semantic structure can complement representations learned primarily from the study sample, even when the host learner is highly flexible.

Third, the gains cannot be explained by generic LLM features alone. Both \textit{LLM} and \textit{LLM+MLP} are substantially weaker than \textsc{CURL}, and in many cases weaker than conventional CATE learners, showing that directly using LLM outputs or embeddings is insufficient. Under the same LLM backbone, \textsc{CURL} also outperforms \textit{GATE} in most settings, indicating that its benefit comes from uncertainty-guided allocation, role-conditioned semantic
construction, and separated routing rather than from raw LLM capacity.

\subsection{Ablation Studies}
\label{subsec:ablation}

\begin{figure}[t]
    \centering
    \includegraphics[width=\columnwidth]{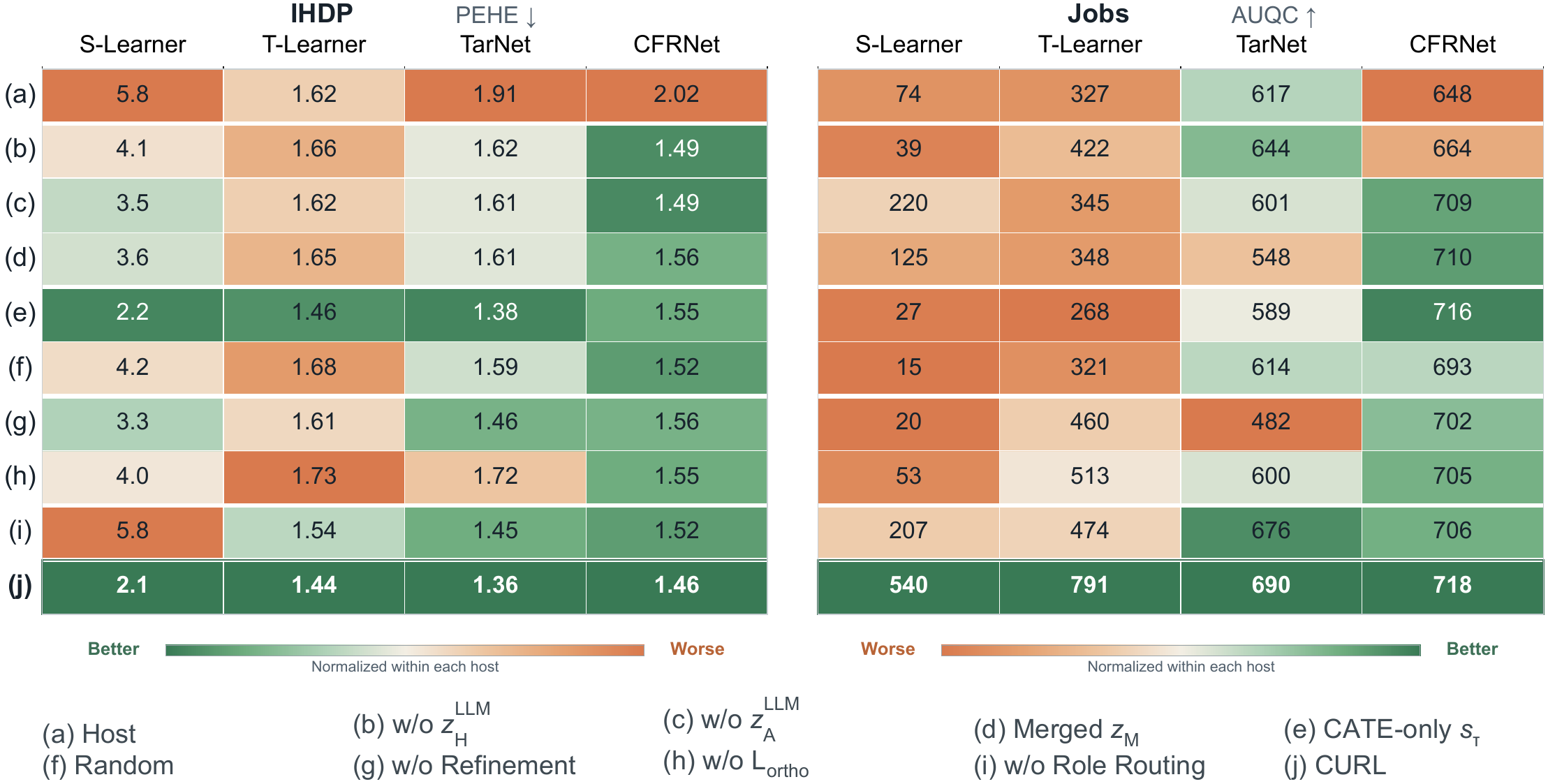}
    \caption{Ablation results across four host learners on IHDP
    (PEHE, lower is better) and Jobs (AUQC, higher is better).
    Each cell reports the mean over five runs. Colors are normalized
    separately within each host-learner column, with greener cells
    indicating better performance.}
    \label{fig:ablation_study}
\end{figure}

Figure~\ref{fig:ablation_study} compares the full \textsc{CURL} with its ablated variants across four representative host learners. The complete model achieves the best result in all eight host--dataset configurations, indicating that its gains do not depend on a particular estimator architecture.

Semantic construction and routing. Removing either semantic channel (\textit{w/o $z_H^{\mathrm{LLM}}$} or \textit{w/o $z_A^{\mathrm{LLM}}$}) consistently underperforms the full model, showing that the two role-conditioned representations provide complementary information. Replacing them with a shared representation (\textit{Merged $z_M$}) also reduces performance. The declines caused by removing $\mathcal L_{\mathrm{ortho}}$ or role-aware routing further support
channel separation and the asymmetric prediction pathways.

Allocation and refinement. Using only CATE uncertainty (\textit{CATE-only $s_\tau$}) remains competitive for several IHDP hosts but loses substantial performance on Jobs, suggesting that assignment-side uncertainty provides an important complementary allocation signal. Random allocation is less reliable and never matches the full uncertainty-guided strategy. Finally, \textit{w/o Refinement} is consistently inferior to \textsc{CURL}, supporting iterative re-diagnosis and semantic reallocation as the host estimator evolves.

\subsection{Sensitivity Analysis}
\label{subsec:sensitivity_analysis}

We assess the robustness of \textsc{CURL} with respect to the LLM backbone, the semantic-allocation ratio $\rho$, and the number of refinement rounds $R$.

\begin{figure}[t]
    \centering
    \includegraphics[width=1\linewidth]{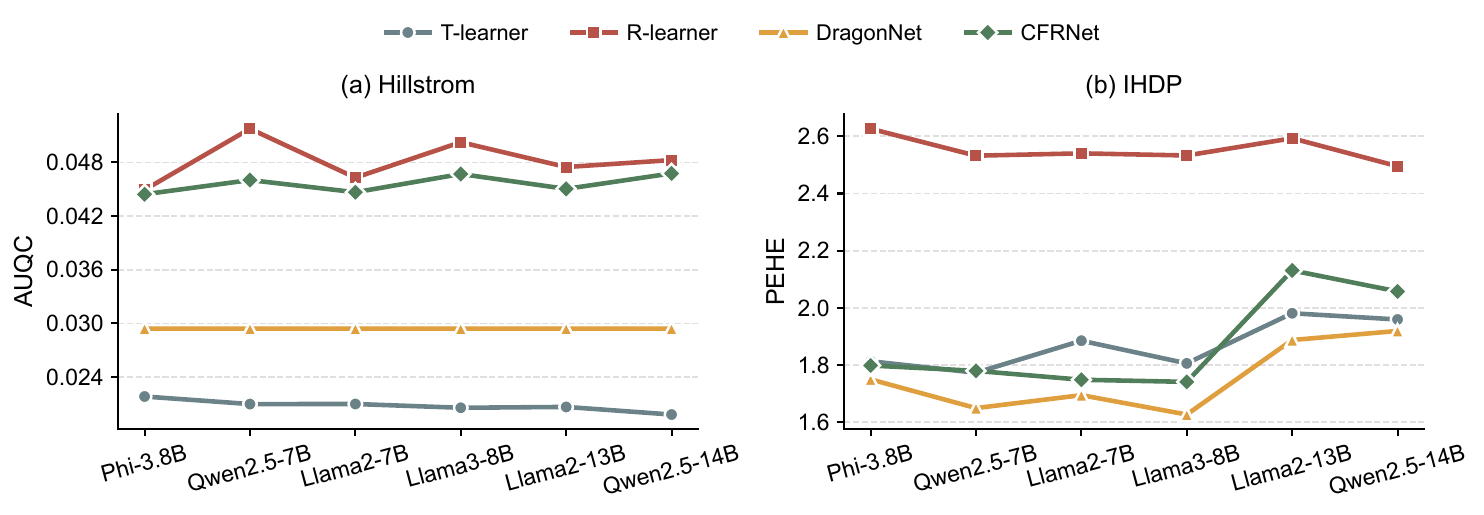}
    \caption{Sensitivity of \textsc{CURL} to different LLM backbones
    on Hillstrom (AUQC) and IHDP (PEHE) across multiple host learners.}
    \label{fig:llm_impact}
\end{figure}

Figure~\ref{fig:llm_impact} reports results under six backbones: Phi-3.8B, Qwen2.5-7B, Llama2-7B, Llama3-8B, Llama2-13B, and Qwen2.5-14B. Performance varies only modestly across backbones, and larger models are not uniformly better. This suggests that \textsc{CURL} is not strongly dependent on a particular backbone or
model scale.

\begin{figure}[t]
    \centering
    \begin{subfigure}[t]{0.475\linewidth}
        \centering
        \includegraphics[width=\linewidth]{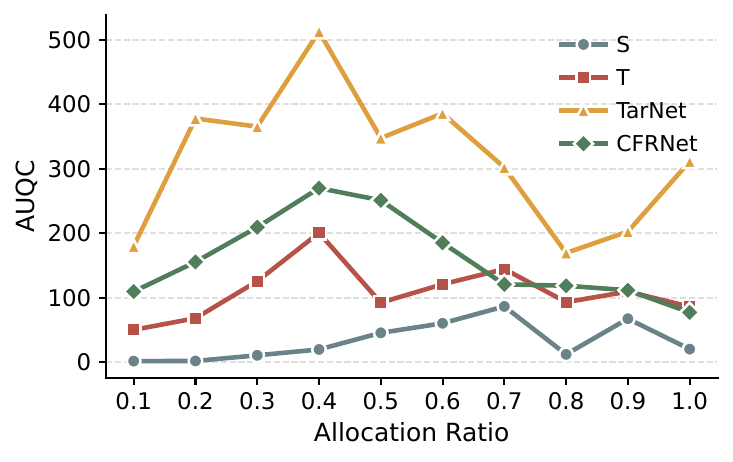}
        \caption{Effect of the allocation ratio.}
        \label{fig:hard_size}
    \end{subfigure}
    \hfill
    \begin{subfigure}[t]{0.475\linewidth}
        \centering
        \includegraphics[width=\linewidth]{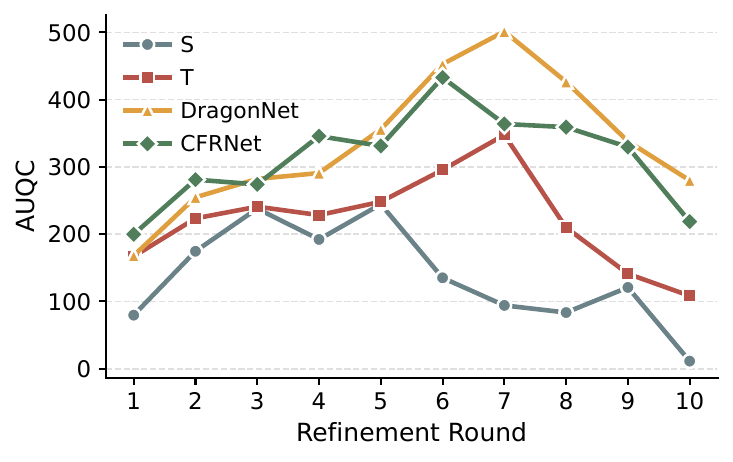}
        \caption{Effect of the refinement rounds.}
        \label{fig:curr_rounds}
    \end{subfigure}
    \caption{Sensitivity of \textsc{CURL} on Jobs (AUQC) across
    multiple host learners: (a) semantic-allocation ratio $\rho$ and
    (b) number of refinement rounds $R$.}
    \label{fig:hardsize_rounds}
\end{figure}

Figure~\ref{fig:hardsize_rounds} examines the allocation ratio $\rho$ and the number of refinement rounds $R$ on Jobs. Performance remains stable over broad ranges, with moderate settings generally performing best. Increasing $\rho$ beyond this range raises the LLM cost without consistent gains, while excessive refinement rounds yield diminishing returns. Overall, \textsc{CURL} is not highly sensitive to either hyperparameter around the selected configuration.

\subsection{Progressive Refinement Dynamics}
\label{subsec:refinement_dynamics}

We further examine how semantic allocation evolves across refinement rounds. Setting $R=5$, we track the cumulative semantic-cache ratio $|\mathcal C^{(r)}|/N$ and the mean uncertainty scores $s_e^{(r)}$ and $s_\tau^{(r)}$ on IHDP and Jobs across four representative host learners. Figure~\ref{fig:refinement_jobs} summarizes the Jobs results in the main text, while the full results are provided in Appendix~\ref{app:refinement_dynamics}.

\begin{figure}[t]
    \centering
    \includegraphics[width=\columnwidth]{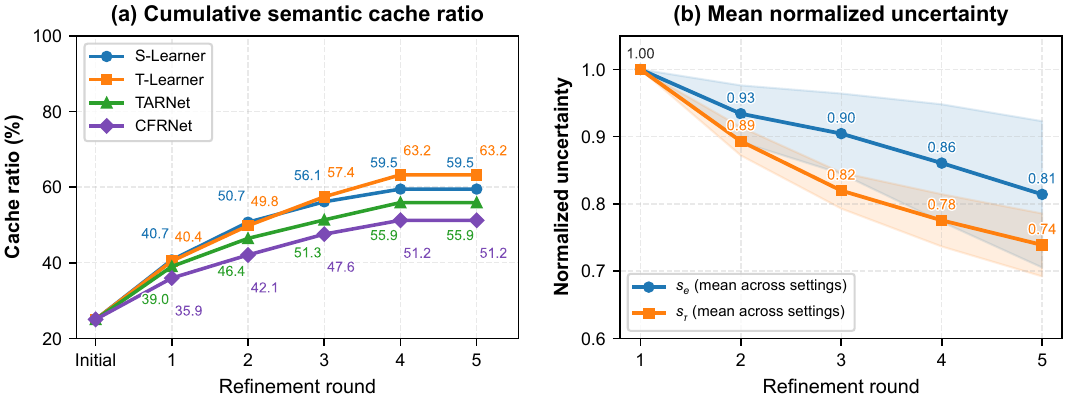}
    \caption{Progressive refinement dynamics on Jobs across four representative host learners. We report the cumulative semantic-cache ratio $|\mathcal C^{(r)}|/N$ and the mean uncertainty scores $s_e^{(r)}$ and $s_\tau^{(r)}$ over refinement rounds.}
    \label{fig:refinement_jobs}
\end{figure}

As shown in Figure~\ref{fig:refinement_jobs}, the semantic cache expands rapidly in the early rounds and then gradually stabilizes. Meanwhile, both uncertainty scores exhibit an overall downward trend across most settings. This pattern is consistent with progressive refinement reducing local predictive instability while the set of units requiring semantic augmentation gradually stabilizes.

\subsection{Role Validation of the Semantic Channels}
\label{subsec:semantic_validation}

We evaluate whether the two LLM-derived channels align with their intended roles through route reassignment and probing, and test whether their gains reflect representational content rather than additional inputs or parameters.

\begin{figure}[t]
    \centering
    \includegraphics[width=\linewidth]
    {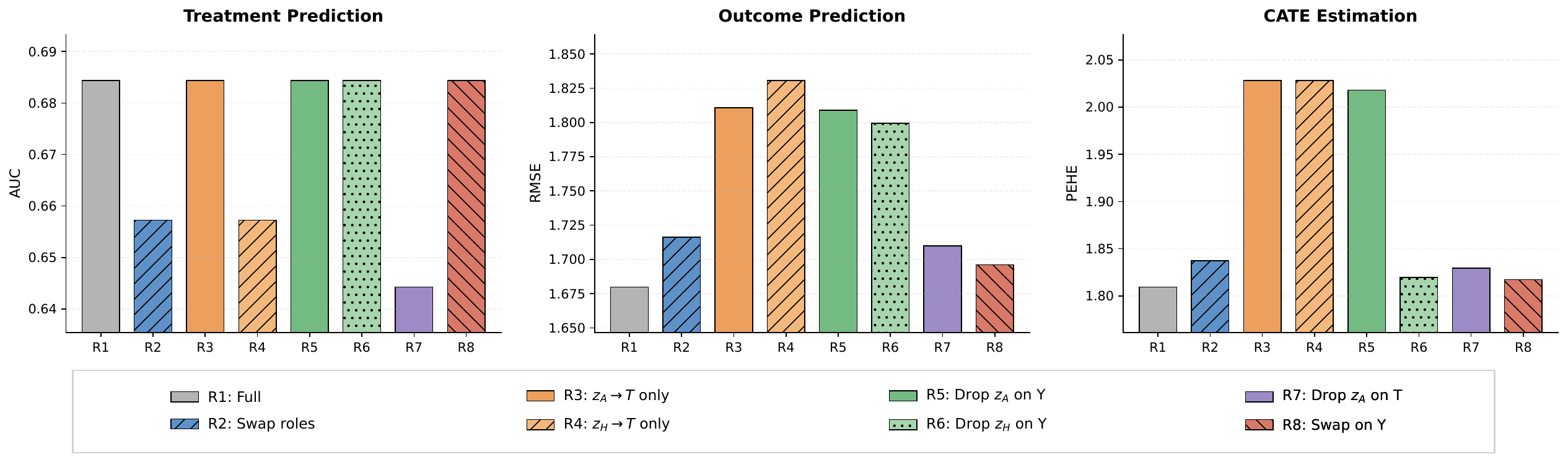}
    \caption{Route-reassignment validation. The original routing performs best overall, while swapping or incorrectly routing the two semantic channels degrades assignment and effect estimation.}
    \label{fig:semantic_route_main}
\end{figure}

Route reassignment. We freeze the trained model and evaluate eight reroutings of $z_A$ and $z_H$ (Figure~\ref{fig:semantic_route_main}). The original routing performs best: swapping the channels or removing $z_A$ from the propensity pathway harms treatment prediction, while incorrect outcome-side routing degrades CATE estimation. This supports the intended assignment and heterogeneity roles.

\begin{figure}[t]
    \centering
    \includegraphics[width=\linewidth]
    {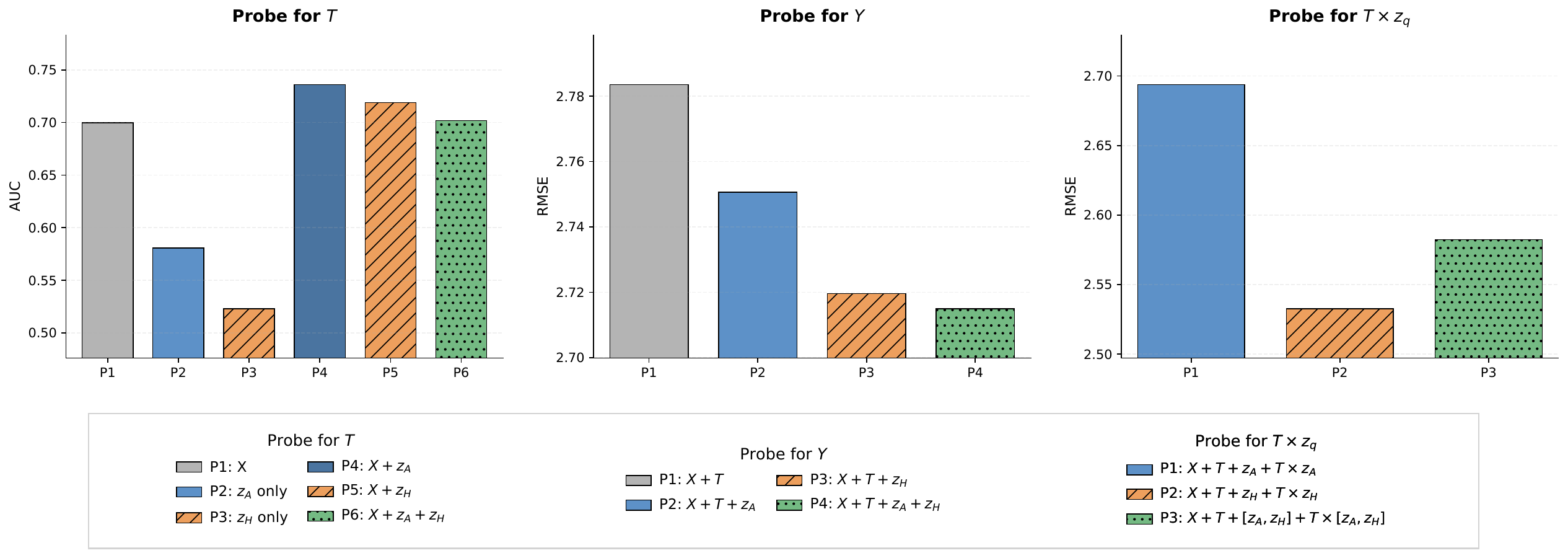}
    \caption{Probe-based role validation. The assignment-oriented
    channel is more informative for treatment prediction, whereas the
    heterogeneity-oriented channel is more informative when modeling
    treatment-dependent responses.}
    \label{fig:semantic_probe}
\end{figure}

Probe analysis. Lightweight probes on frozen representations provide consistent evidence (Figure~\ref{fig:semantic_probe}): $z_A$ better predicts treatment assignment, whereas $z_H$ better predicts treatment-dependent responses when treatment--representation interactions are included. Combining both channels performs best, indicating complementary information.

Representational-lossiness checks. Semantic-corruption results on IHDP (Appendix~\ref{app:representation_lossiness}) show that CURL performs best: anonymizing feature names and shuffling category labels both reduce its advantage, while random embeddings cause the largest degradation. This ordering indicates that CURL's improvement derives in part from semantic structure encoded by the LLM, rather than additional adapter capacity alone; Appendix~\ref{app:case_study} illustrates the channels' content.

\section{Conclusion}

We presented CURL, a plug-in framework for finite-sample CATE estimation under representational lossiness, where causally sufficient covariates remain difficult to exploit in their raw encoded form. Rather than using LLMs as unstructured feature generators or direct outcome imputers, \textsc{CURL} uses estimator uncertainty to identify locally unstable units, constructs assignment- and heterogeneity-oriented representations from their observed covariates, and integrates them through role-aware pathways. Across four benchmarks and ten host learners, \textsc{CURL} improves performance in most settings, while ablation, refinement, corruption, and role-validation analyses support uncertainty-guided allocation and channel separation. \textsc{CURL} reorganizes observed covariates rather than introducing new unit-level information; its generated text is a semantic representation and does not address unmeasured confounding or overlap violations. Future work should examine semantic reliability and uncertainty calibration under domain shift. Overall, \textsc{CURL} shows how a pretrained semantic prior can support more reliable finite-sample CATE estimation by making these relations easier to exploit, without changing the observed information set, target estimand, or identifying assumptions.
\bibliographystyle{plainnat}
\bibliography{refs}

\newpage
\appendix

\section{Dataset and Implementation Details}
\label{app:exp_details}

This appendix provides additional benchmark, implementation, and reproducibility details omitted from Section~\ref{subsec:experimental_setup}.

\subsection{Benchmark Details}
\label{app:benchmark_details}

\paragraph{IHDP}
IHDP is a semi-synthetic benchmark based on covariates from the Infant Health and Development Program. Treatment assignment and potential outcomes follow the standard semi-synthetic construction, providing ground-truth individual treatment effect for evaluation. The covariates describe infant health, maternal background, pregnancy risk factors, and prenatal context.

\paragraph{Adult.}
Adult is constructed from the Adult income dataset with a semi-synthetic treatment and outcome mechanism. The treatment represents a career-opportunity exposure related to observed demographic and employment characteristics, while the outcome indicates high-income status. The simulator provides ground-truth individual effects.

\paragraph{Jobs.}
Jobs evaluates the effect of participation in a job-training program. The observed covariates include demographic characteristics, education, and pre-treatment earnings histories. Since individual-level counterfactual outcomes are unavailable, we evaluate the quality of the estimated treatment-effect ranking using uplift metrics.

\paragraph{Hillstrom.}
Hillstrom is an original randomized email-marketing experiment. We compare the Men's E-Mail treatment with the no-email control. Its covariates describe customer tenure, recency, historical spending, purchase preferences, transaction channel, and geographic segment. As individual-level effects are unavailable, evaluation uses uplift metrics.

\subsection{Additional Implementation Details}
\label{app:exp_details_impl}

\paragraph{Baselines.}
We plug \textsc{CURL} into ten host learners spanning three families: meta-learners (S-, T-, and R-Learner), balanced-representation methods (TARNet, CFRNet, and DragonNet), and deep latent-variable methods (CEVAE, TEDVAE, SDD, and CiVAE). We additionally include two LLM-only baselines. \textit{LLM} prompts the LLM to directly predict CATE from the textualized profile, while \textit{LLM+MLP} uses the frozen LLM as an encoder and trains an MLP prediction head. \textit{GATE} \citep{huynh2025improving} directly generates counterfactual outcomes from selected covariates. Additional comparisons with standalone feature-disentanglement methods are reported in
Appendix~\ref{app:additional_baseline}.

\paragraph{Semantic construction.}
For every unit in the newly surfaced query set $\mathcal Q^{(r)}$, we issue two independent prompts, $P_{A,i}$ and $P_{H,i}$. They construct assignment-oriented and heterogeneity-oriented representations, respectively, from the same observed profile. The two prompts use separate dialogue states and key-value caches. Complete prompt, profile, and $\texttt{insight\_hint}$ templates are given in Appendix~\ref{app:prompts}.

\paragraph{Hyperparameters.}
The main experiments use $M=30$ MC-dropout passes, semantic-allocation ratio $\rho=0.25$, and progressive refinement with $R=3$, unless otherwise stated. The objective weights are $\lambda_{\mathrm{prop}}=1.0$ and $\lambda_{\mathrm{ortho}}=0.1$. Qwen2.5-7B or Llama3-8B is selected
on the validation set for each benchmark; additional backbones are
examined in Section~\ref{subsec:sensitivity_analysis}.

\paragraph{Optimization and data splits.}
For each benchmark, 80\% of the samples are allocated to the training partition, with a validation subset used for host-model selection, LLM-backbone selection, hyperparameter selection, and early stopping. All methods use identical splits under each random seed.

\paragraph{Hardware and reproducibility.}
Experiments are run on two GPUs with 48\,GB memory each. Main results are reported over five random seeds.

\subsection{Additional Comparison with Feature-Disentanglement Baselines}
\label{app:additional_baseline}

We further compare against three standalone feature-disentanglement
baselines: DR-CFR~\citep{hassanpour2020learning},
DeR-CFR~\citep{wu2023learning}, and
DDRN~\citep{meng2025deep}. These methods learn decomposed
representations from the observed sample and are trained as independent
estimators rather than \textsc{CURL} hosts. To keep the comparison
compact and consistent across metrics, we report a fixed
\textsc{CURL}-enhanced R-Learner, denoted by \textsc{CURL-R}, on Jobs
and Adult.

Tables~\ref{tab:disentangle_jobs}
and~\ref{tab:disentangle_adult} show that \textsc{CURL-R} achieves
the best result on all reported metrics. This suggests that pretrained
semantic structure provides an inductive bias complementary to
feature disentanglement learned solely from the study sample.

\begin{table}[t]
\centering
\caption{Additional comparison with feature-disentanglement baselines
on Jobs. Higher values are better.}
\label{tab:disentangle_jobs}
\setlength{\tabcolsep}{5.0pt}
\renewcommand{\arraystretch}{1.08}
\begin{tabular}{lccc}
\toprule
Method
& AUUC $\uparrow$
& AUQC $\uparrow$
& LIFT@30 $\uparrow$ \\
\midrule
LLM
& 336.07
& 51.52
& 3.25 \\
LLM+MLP
& 810.99
& 69.11
& 2.56 \\
DR-CFR
& 745.90
& 206.14
& 7.93 \\
DeR-CFR
& 774.31
& 209.34
& 7.68 \\
DDRN
& 734.14
& 228.01
& 7.63 \\
\textsc{CURL-R}
& \textbf{1007.40}
& \textbf{240.12}
& \textbf{8.14} \\
\bottomrule
\end{tabular}
\end{table}

\begin{table}[t]
\centering
\caption{Additional comparison with feature-disentanglement baselines
on Adult. Lower values are better.}
\label{tab:disentangle_adult}
\setlength{\tabcolsep}{5.0pt}
\renewcommand{\arraystretch}{1.08}
\begin{tabular}{lccc}
\toprule
Method
& PEHE $\downarrow$
& $\epsilon_{\mathrm{ATE}}$ $\downarrow$
& $\epsilon_{\mathrm{ATT}}$ $\downarrow$ \\
\midrule
LLM
& 57.312
& 1.569
& 0.708 \\
LLM+MLP
& 0.429
& 0.191
& 0.170 \\
DR-CFR
& 0.092
& 0.010
& 0.014 \\
DeR-CFR
& 0.089
& 0.010
& 0.015 \\
DDRN
& 0.094
& 0.022
& 0.012 \\
\textsc{CURL-R}
& \textbf{0.062}
& \textbf{0.004}
& \textbf{0.002} \\
\bottomrule
\end{tabular}
\end{table}

\section{Detailed Definitions of Evaluation Metrics}
\label{app:metrics}

\paragraph{Uplift metrics (Jobs and Hillstrom).}
Individual-level counterfactual outcomes are unavailable on Jobs and
Hillstrom, so pointwise treatment-effect errors cannot be computed.
We instead evaluate the ranking induced by
$\widehat{\tau}(x)$. Let $i_{(1)},\ldots,i_{(N)}$ denote the units
sorted in descending order of $\widehat{\tau}(x_i)$, and let
\[
\pi(k)=\{i_{(1)},\ldots,i_{(k)}\}
\]
be the top-$k$ ranked units. For $a\in\{0,1\}$, define the cumulative
outcome and sample count within the top-$k$ prefix as
\begin{align}
S_a(k)
&=
\sum_{\substack{i\in\pi(k)\\T_i=a}}Y_i,
&
N_a(k)
&=
\sum_{i\in\pi(k)}\mathbb{I}(T_i=a).
\end{align}
We use the convention
\begin{equation}
\overline{Y}_a(k)
=
\begin{cases}
S_a(k)/N_a(k), & N_a(k)>0,\\
0,             & N_a(k)=0.
\end{cases}
\end{equation}

The cumulative uplift-curve ordinate at cutoff $k$ is
\begin{equation}
U(k)
=
k\left(
\overline{Y}_1(k)-\overline{Y}_0(k)
\right).
\end{equation}
Let $x_k=k/N$. We report the sample-size-normalized trapezoidal
area under this curve:
\begin{align}
\mathrm{AUUC}
&=
\frac{1}{N}
\operatorname{Trapz}
\left(
\left\{(x_k,U(k))\right\}_{k=1}^{N}
\right)
\\
&=
\frac{1}{2N^2}
\sum_{k=1}^{N-1}
\left[
U(k)+U(k+1)
\right].
\end{align}
The additional factor $1/N$ reduces the dependence of the reported
value on the evaluation-set size.

For the Qini curve, let
\begin{equation}
N_T=N_1(N),
\qquad
N_C=N_0(N),
\end{equation}
and assume $N_T>0$ and $N_C>0$. The Qini ordinate uses the
full-sample treated-to-control count ratio:
\begin{equation}
Q(k)
=
S_1(k)
-
\frac{N_T}{N_C}S_0(k).
\end{equation}
The reported normalized area under the Qini curve is
\begin{align}
\mathrm{AUQC}
&=
\frac{1}{N}
\operatorname{Trapz}
\left(
\left\{(x_k,Q(k))\right\}_{k=1}^{N}
\right)
\\
&=
\frac{1}{2N^2}
\sum_{k=1}^{N-1}
\left[
Q(k)+Q(k+1)
\right].
\end{align}
If either treatment arm is absent from the evaluation set, AUQC is
defined as zero.

Finally, define the empirical uplift within a ranked prefix as
\begin{equation}
L(k)
=
\begin{cases}
\dfrac{S_1(k)}{N_1(k)}
-
\dfrac{S_0(k)}{N_0(k)},
&
N_1(k)>0\ \text{and}\ N_0(k)>0,
\\[6pt]
0, & \text{otherwise}.
\end{cases}
\end{equation}
Let
\begin{equation}
K_{30}=\lfloor0.3N\rfloor
\end{equation}
and define the full-sample empirical uplift as
\begin{equation}
L_{\mathrm{all}}
=
\frac{S_1(N)}{N_T}
-
\frac{S_0(N)}{N_C}.
\end{equation}
We report the relative top-30\% lift
\begin{equation}
\mathrm{LIFT@30}
=
\frac{L(K_{30})}
     {L_{\mathrm{all}}+\varepsilon},
\qquad
\varepsilon=10^{-9}.
\end{equation}
Thus, LIFT@30 is dimensionless and measures the uplift in the
highest-ranked 30\% of units relative to the overall empirical
uplift. AUUC, AUQC, and LIFT@30 are higher-is-better under the
evaluation protocol used in our experiments.
\section{Prompt Construction and Semantic Extraction}
\label{app:prompts}

This appendix specifies the construction of the dataset-level
$\texttt{insight\_hint}$, the two role-conditioned prompts
$P_A$ and $P_H$, the deterministic profile constructors, and the
semantic-embedding extraction procedure.

\subsection{Insight Hint}
\label{app:insight_hint}

At initialization, we construct the dataset-level $\texttt{insight\_hint}$ once from the training set by comparing the initially selected units $\mathcal S^{(0)}$ with the full training population using only observed covariates. For each numeric field, we compute the relative difference between the two means and retain at most ten fields whose absolute difference exceeds $20\%$, indicating whether each field is higher or lower in $\mathcal S^{(0)}$. This aggregate summary is inserted into the template below, from which the LLM generates a one-sentence hint $h_{\mathrm{ins}}$. The hint is then reused unchanged for that dataset throughout all refinement rounds and contains no unit-level treatment assignments, outcomes, counterfactuals, or test-set information.

\begin{tcblisting}{
  colback=gray!5,
  colframe=gray!50,
  title={Template for \texttt{insight\_hint}},
  fonttitle=\bfseries\small,
  listing only,
  listing options={
    basicstyle=\small\ttfamily,
    breaklines=true,
    columns=fullflexible
  },
  breakable
}
[System] You are a Causal Inference Researcher.

[Context] The scenario is about marketing / medical / other. The statistical model fails to predict accurately for a specific subgroup.

[Hard Subgroup Characteristics] {stat_summary}

[Task] Provide a brief, high-level hypothesis about why the estimator failed.
\end{tcblisting}

\subsection{Role-Conditioned Prompt Templates}
\label{app:prompt-template}

For every queried unit $i$, the deterministic profile
$p_i=\psi(x_i)$ and fixed hint $h_{\mathrm{ins}}$ are inserted into
two independent prompts. The assignment-oriented prompt
$P_{A,i}$ focuses on pre-treatment baseline structure relevant to the
shared and assignment-side components of the estimator. The
heterogeneity-oriented prompt $P_{H,i}$ focuses on stable
pre-treatment attributes relevant to treatment-response variation.
Both prompts explicitly restrict the LLM to semantic relations
supported by the observed profile.

\paragraph{Assignment-oriented prompt $P_A$.}

\begin{tcblisting}{
  colback=gray!5,
  colframe=gray!50,
  title={$\mathrm{Prompt}_A$: Pre-treatment Assignment-Side Inference},
  fonttitle=\bfseries\small,
  listing only,
  listing options={
    basicstyle=\small\ttfamily,
    breaklines=true,
    columns=fullflexible
  },
  breakable
}
[System] You are a Causal Analyst specializing in confounding bias.

[Constraint] TEMPORAL FIREWALL ACTIVATED.
You are strictly limited to the Pre-Treatment Timeline.
You must IGNORE any events that happen after the intervention/contact.
You must IGNORE the final outcome (Y).

[Background Insight] {insight_hint}
[Subject Profile] {Prof_i}

[Task] Analyze the baseline characteristics that existed BEFORE any action was taken. Based only on the observed covariates X, identify semantic relations and interactions that may be relevant to modeling treatment assignment and baseline outcome patterns.

Such as:
1. Socio-economic constraints rooted in the past.
2. Fundamental health/financial status established prior to this event.
3. Historical behavioral patterns.

[Goal] Compress these pre-existing confounders into a single representation.

[Trigger] Pre-treatment Confounding Context:
\end{tcblisting}

\paragraph{Heterogeneity-oriented prompt $P_H$ for IHDP, Jobs, and Hillstrom.}

\begin{tcblisting}{
  colback=gray!5,
  colframe=gray!50,
  title={$\mathrm{Prompt}_H$: General Response-Side Heterogeneity Inference},
  fonttitle=\bfseries\small,
  listing only,
  listing options={
    basicstyle=\small\ttfamily,
    breaklines=true,
    columns=fullflexible
  },
  breakable
}
[System] You are a Causal Analyst specializing in effect heterogeneity.

[Constraint] ATTRIBUTE STABILITY CHECK. Focus ONLY on stable, inherent traits of the subject that determine their sensitivity.

[Background Insight] {insight_hint}
[Subject Profile] {Prof_i}

[Task] Analyze the inherent modifiers that dictate how the subject responds to potential treatments.

Such as:
1. Psychological resilience or sensitivity (Stable Personality).
2. Biological or structural receptivity.
3. Risk tolerance levels.

[Goal] Compress these heterogeneity modifiers into a single representation.

[Trigger] Effect Modification Context:
\end{tcblisting}

\paragraph{Specialized $P_H$ for Adult.}

Adult uses a domain-specific variant because its simulated heterogeneity is primarily associated with employment structure, education, and socio-economic mobility.

\begin{tcblisting}{
  colback=gray!5,
  colframe=gray!50,
  title={$\mathrm{Prompt}_H$ for Adult: Socio-Economic Modifier Inference},
  fonttitle=\bfseries\small,
  listing only,
  listing options={
    basicstyle=\small\ttfamily,
    breaklines=true,
    columns=fullflexible
  },
  breakable
}
[System] You are a Sociological & Economic Analyst.

[Constraint] ATTRIBUTE STABILITY CHECK. Focus ONLY on socio-economic traits.

[Background Insight] {insight_hint}
[Subject Profile] {Prof_i}

[Task] Analyze the inherent modifiers that dictate how this individual responds to career or income interventions (e.g., job training).

Such as:
1. Social mobility potential based on their current class and education.
2. Human capital accumulation potential.

[Goal] Compress these heterogeneity modifiers into a single representation.

[Trigger] Socio-economic Effect Modification Context
\end{tcblisting}

\paragraph{Semantic extraction.}
For each prompt, the frozen LLM first generates a deterministic textual
response. We then perform a second forward pass over the concatenated
prompt and generated response and extract the final-layer hidden state
of the last valid generated-response token. This yields
\begin{equation}
z_{A,i}^{\mathrm{LLM}}
=
\operatorname{Emb}_{\Phi}(P_{A,i}),
\qquad
z_{H,i}^{\mathrm{LLM}}
=
\operatorname{Emb}_{\Phi}(P_{H,i}),
\end{equation}
consistent with Eq.~\eqref{eq:emb}. The generated text is used only
for inspection; the adapter consumes the corresponding hidden-state
embedding.

\subsection{Dataset-Specific Profile Templates}
\label{app:prompt-instances}

The profile constructor $\psi$ deterministically renders each
structured covariate vector into natural language. No treatment
assignment, factual outcome, counterfactual outcome, prediction error,
or test statistic is included. A dataset-level treatment definition is
included only to specify the meaning of the intervention.

\paragraph{IHDP profile.}

\begin{tcblisting}{
  colback=gray!5,
  colframe=gray!50,
  title={IHDP profile template},
  fonttitle=\bfseries\small,
  listing only,
  listing options={
    basicstyle=\small\ttfamily,
    breaklines=true,
    columns=fullflexible
  },
  breakable
}
Child Profile: A {sex} infant born {weeks_preterm} weeks preterm with a birth weight of {birth_weight} kg. Neonatal health index is recorded at {neonatal_health_index}.

Maternal Background: The mother was {mother_age} years old and {marital_status} at the time of birth. Her education level is {education_level}.

Pregnancy History: {pregnancy_risk_statement}

Context: The mother {prenatal_care_status} prenatal care. She was {work_status} during the study period.

Treatment Definition: Treatment means receiving the IHDP early-childhood intervention program; control means not receiving that program.
\end{tcblisting}

Here, the pregnancy-risk statement is formed deterministically from
the observed smoking, alcohol-use, and drug-use indicators.

\paragraph{Adult profile.}

\begin{tcblisting}{
  colback=gray!5,
  colframe=gray!50,
  title={Adult profile template},
  fonttitle=\bfseries\small,
  listing only,
  listing options={
    basicstyle=\small\ttfamily,
    breaklines=true,
    columns=fullflexible
  },
  breakable
}
Demographics: A {age}-year-old {gender} ({race}) from {native_country}.

Socio-economic: Works as a {occupation} in the {workclass} sector, with {education} education level. Works {hours_per_week} hours per week.

Family Status: Currently {marital_status}, family role is {relationship}.

Treatment Definition: Treatment is a semi-synthetic career-opportunity exposure.
\end{tcblisting}

\paragraph{Jobs profile.}

\begin{tcblisting}{
  colback=gray!5,
  colframe=gray!50,
  title={Jobs profile template},
  fonttitle=\bfseries\small,
  listing only,
  listing options={
    basicstyle=\small\ttfamily,
    breaklines=true,
    columns=fullflexible
  },
  breakable
}
Subject Profile: A {age}-year-old {race} {marital_status} individual, {degree_status}, with {years_of_education} years of education.

Pre-treatment Financials: Real earnings in 1974 were ${earnings_1974}, and in 1975 were ${earnings_1975}.

Treatment Definition: Treatment means participating in the job-training program; control means not participating.
\end{tcblisting}

\paragraph{Hillstrom profile.}

\begin{tcblisting}{
  colback=gray!5,
  colframe=gray!50,
  title={Hillstrom profile template},
  fonttitle=\bfseries\small,
  listing only,
  listing options={
    basicstyle=\small\ttfamily,
    breaklines=true,
    columns=fullflexible
  },
  breakable
}
Customer Persona: This individual {customer_status}.

Engagement History: They last interacted with the brand {recency} months ago.

Shopping Preferences: They {mens_purchase_history} and {womens_purchase_history}.

Financial Profile: Their total historical value is ${history}, officially categorized in the '{history_segment}' segment.

Behavioral Context: They primarily use the '{channel}' channel for transactions and are located in a {zip_code_type} environment.

Treatment Definition: Treatment means receiving the Men's E-Mail campaign; control means receiving no e-mail campaign.
\end{tcblisting}

\section{Training Algorithm}
\label{app:algorithm}

Algorithm~\ref{alg:curl} summarizes training. We use $R$ to denote
the total number of semantic-assisted progressive optimization rounds,
including the first round initialized by the initial semantic
allocation. Accordingly, the rounds are indexed by
$r=0,\ldots,R-1$, and the final wrapped estimator is $F^{(R)}$.

The initial allocation at $r=0$ uses the same joint ranking rule as
the subsequent rounds, but $s_{e,i}^{(0)}$ is uniformly set to zero.
Hence, the initial ordering reduces to the ordering induced by
$s_{\tau,i}^{(0)}$ without introducing a separate allocation rule.
After the first optimization round, every wrapped host contains a
trained propensity head. Therefore, both $s_{e,i}^{(r)}$ and
$s_{\tau,i}^{(r)}$ are used in the remaining refinement rounds
$r=1,\ldots,R-1$.

The selected set $\mathcal S^{(r)}$, cumulative semantic cache
$\mathcal C^{(r)}$, and newly surfaced query set
$\mathcal Q^{(r)}$ follow Eq.~\eqref{eq:cache_update}. The LLM is
queried only for $\mathcal Q^{(r)}$, so every training unit is queried
at most once.

\begin{algorithm}[t]
\small
\DontPrintSemicolon
\caption{CURL training with progressive refinement}
\label{alg:curl}

\KwIn{
dataset $\mathcal D$;
host learner;
frozen LLM $\Phi$;
$\rho,M,R,\lambda_{\mathrm{prop}},
\lambda_{\mathrm{ortho}}$
}
\KwOut{trained wrapped estimator $F^\star$}

Fit the unaugmented host using its native objective\;

Initialize the wrapped estimator $F^{(0)}$ with zero semantic inputs\;

$\mathcal C^{(-1)}\leftarrow\varnothing$\;

\BlankLine
\For{$r=0,\ldots,R-1$}{

    \eIf{$r=0$}{
        \tcc{Initial semantic allocation}
        Run $M$ MC-dropout passes of $F^{(0)}$ and compute
        $s_{\tau,i}^{(0)}$ for all $i$\;

        Set $s_{e,i}^{(0)}\leftarrow0$ for all $i$\;
    }{
        \tcc{Refinement allocation}
        Run $M$ MC-dropout passes of $F^{(r)}$\;

        Compute $s_{e,i}^{(r)}$ and $s_{\tau,i}^{(r)}$ using
        Eq.~\eqref{eq:sig_defs}\;
    }

    Form $\mathcal S^{(r)}$ using Eq.~\eqref{eq:hard_iter}\;

    $\mathcal C^{(r)}
    \leftarrow
    \mathcal C^{(r-1)}\cup\mathcal S^{(r)}$\;

    $\mathcal Q^{(r)}
    \leftarrow
    \mathcal S^{(r)}\setminus\mathcal C^{(r-1)}$\;

    Query the LLM only for units in $\mathcal Q^{(r)}$ to obtain
    $z_{A,i}^{\mathrm{LLM}}$ and
    $z_{H,i}^{\mathrm{LLM}}$\;

    Construct $z_{A,i}^{(r)}$ and $z_{H,i}^{(r)}$ using
    Eq.~\eqref{eq:sparse_mem}\;

    Optimize the host, projectors, gates, and propensity head from
    the current parameters using Eq.~\eqref{eq:overall_loss},
    obtaining $F^{(r+1)}$\;
}

$F^\star\leftarrow F^{(R)}$\;

\Return{$F^\star$}\;
\end{algorithm}
\section{Adapter Instantiation Across Host Learners}
\label{appendix:adapters}

For every host, the attached propensity head consumes the assignment-adjusted representation $u_A$, while the host-specific effect operator consumes $u_H$, which already contains $u_A$ together with the heterogeneity-oriented feature block. The host's native parameterization and main objective are otherwise retained.

\begin{table*}[t]
\centering
\caption{Instantiation of \textsc{CURL} across the ten host learners.
The propensity component consumes $u_A$, while the effect component
consumes $u_H$. The complete objective is
$\mathcal L_{\mathrm{main}}
+\lambda_{\mathrm{prop}}\mathcal L_{\mathrm{prop}}
+\lambda_{\mathrm{ortho}}\mathcal L_{\mathrm{ortho}}$.}
\label{tab:adapters}
\setlength{\tabcolsep}{4pt}
\renewcommand{\arraystretch}{1.15}
\resizebox{\textwidth}{!}{
\begin{tabular}{@{}lllll@{}}
\toprule
\textbf{Pattern}
& \textbf{Host learner(s)}
& \textbf{Assignment/shared pathway}
& \textbf{Effect pathway}
& \textbf{Native main objective} \\
\midrule

Treatment-conditioned outcome
& S-Learner
& $\widehat e\leftarrow u_A$
& $\widehat\mu(u_H,t)$
& Factual outcome loss \\

\midrule

Separate potential-outcome heads
& T-Learner
& $\widehat e\leftarrow u_A$
& $\widehat\mu_0(u_H),\widehat\mu_1(u_H)$
& Arm-specific factual losses \\

\midrule

Shared representation with outcome heads
& TARNet, CFRNet, DragonNet
& $\widehat e\leftarrow u_A$
& $\widehat\mu_0(u_H),\widehat\mu_1(u_H)$
& Native factual, balancing, or targeted loss \\

\midrule

Direct CATE
& R-Learner
& $\widehat e,\widehat\mu\leftarrow u_A$
& $\widehat\tau\leftarrow u_H$
& Orthogonalized R-loss~\citep{nie2021quasi} \\

\midrule

Latent-variable
& CEVAE, TEDVAE, SDD, CiVAE
& Propensity and shared baseline components use $u_A$
& Outcome/CATE operator additionally conditions on $u_H$
& Native ELBO or disentanglement objective \\

\bottomrule
\end{tabular}
}
\end{table*}

\begin{figure*}[t]
    \centering
    \includegraphics[width=0.8\textwidth]{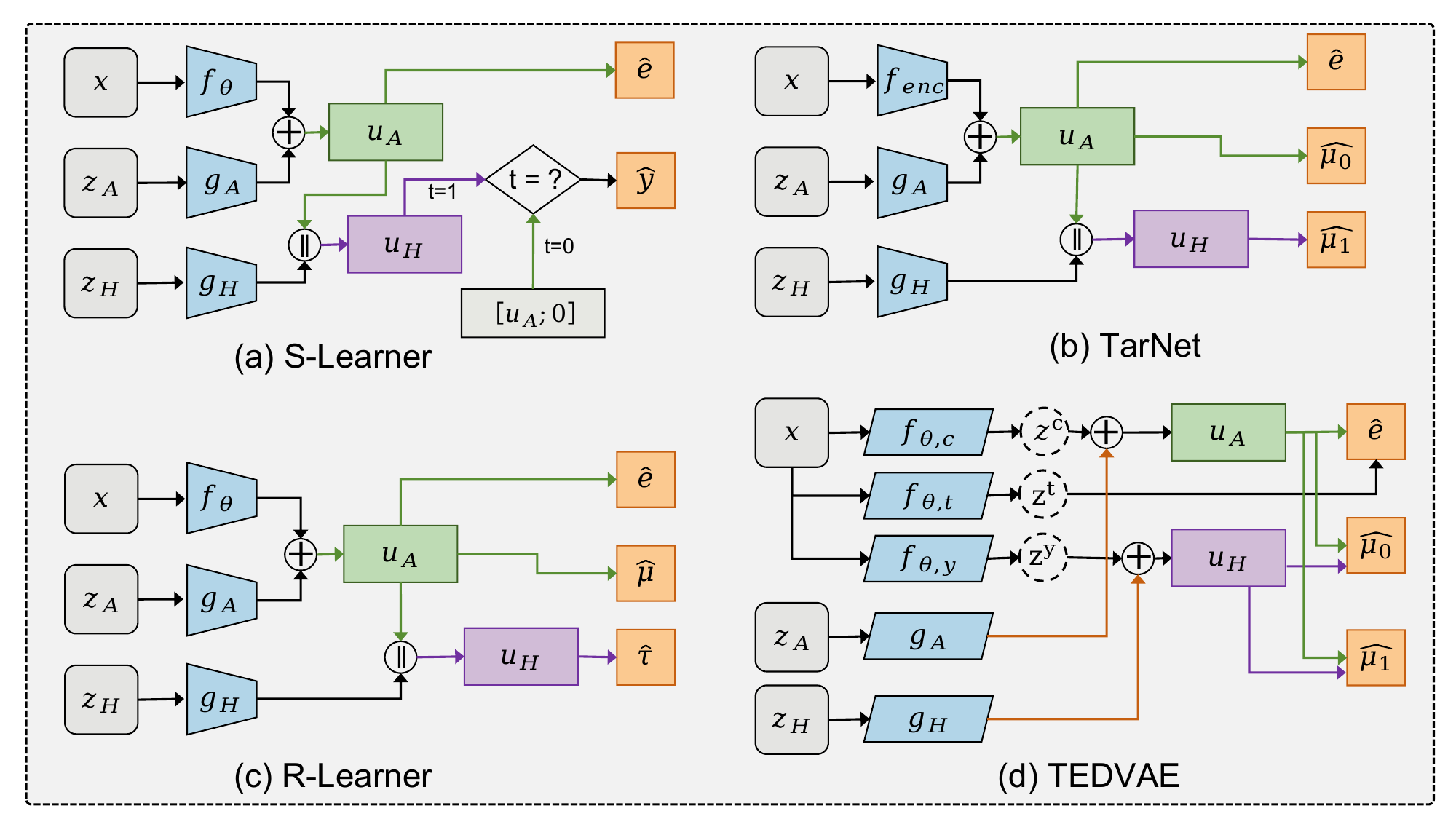}
    \caption{Representative \textsc{CURL} instantiations for
    (a) S-Learner, (b) TARNet, (c) R-Learner, and (d) TEDVAE.
    Across hosts, the propensity pathway consumes $u_A$, whereas the
    host-specific effect operator consumes $u_H$. The insertion point
    varies with the host architecture, while the role-aware interface
    remains unchanged.}
    \label{fig:fusion_instance}
\end{figure*}

Table~\ref{tab:adapters} summarizes the common interface rather than claiming an identifiable causal decomposition of the host's internal variables. The assignment-oriented representation contributes to the shared state used by the propensity component, while the heterogeneity-oriented representation is retained as a separate block available only to the effect operator. This interface accommodates treatment-conditioned predictors, separate potential-outcome heads, direct-CATE learners, and latent-variable estimators.

\section{Additional Analysis of Progressive Refinement}
\label{app:refinement_dynamics}

We perform an initial semantic allocation followed by five refinement
rounds on IHDP and Jobs using S-Learner, T-Learner, TARNet, and
CFRNet. For the cache analysis, define
\begin{equation}
\kappa^{(r)}
=
\frac{|\mathcal C^{(r)}|}{N}.
\end{equation}
For each refinement round, we also report the mean uncertainty scores
\begin{equation}
\overline{s}_{e}^{(r)}
=
\frac{1}{N}
\sum_{i=1}^{N}s_{e,i}^{(r)},
\qquad
\overline{s}_{\tau}^{(r)}
=
\frac{1}{N}
\sum_{i=1}^{N}s_{\tau,i}^{(r)}.
\end{equation}

\begin{figure*}[t]
    \centering
    \includegraphics[width=\textwidth]
    {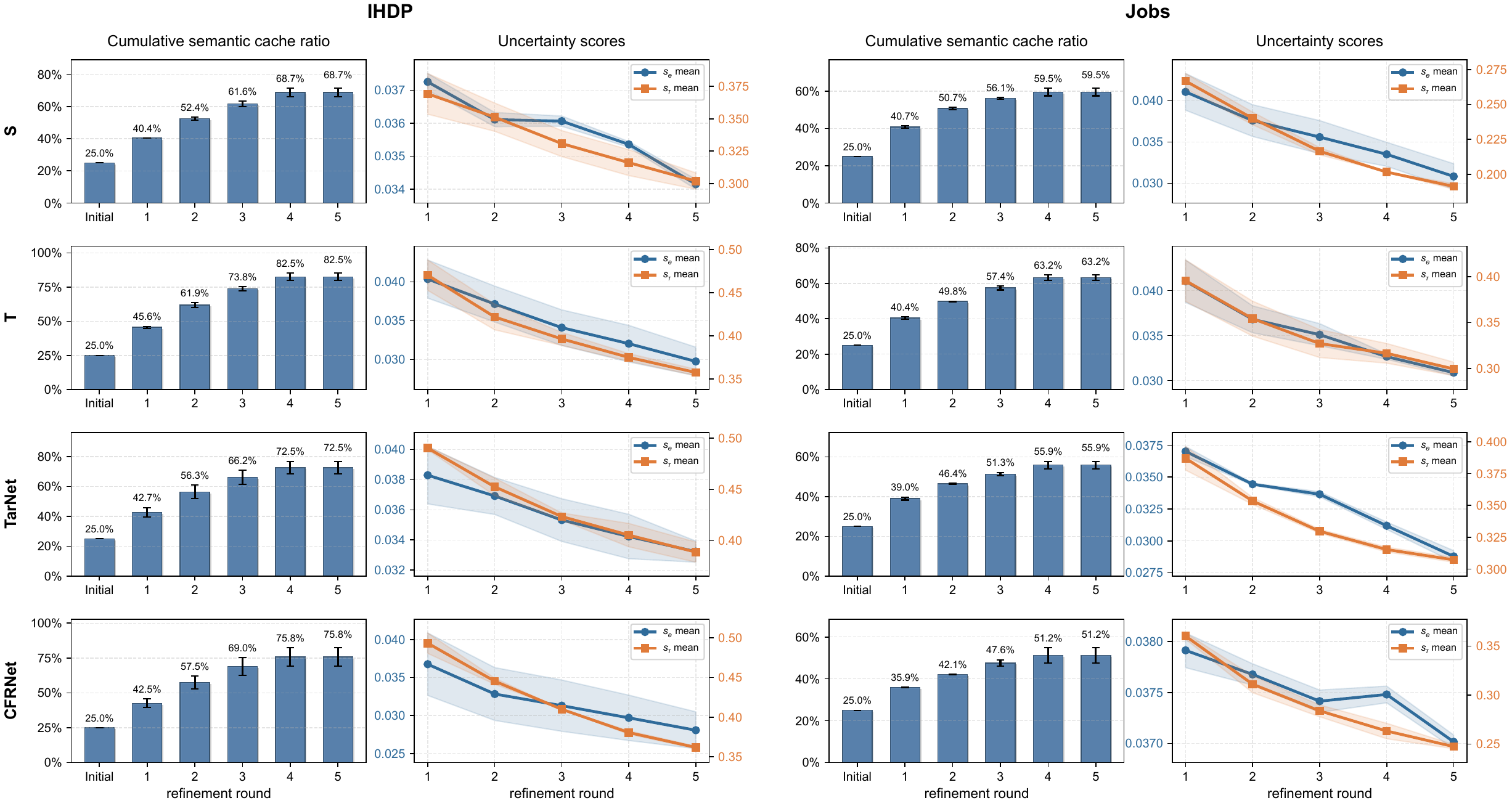}
    \caption{Progressive-refinement dynamics on IHDP and Jobs.
    For each host, the left panel reports the cumulative
    semantic-cache ratio from the initial allocation through five
    refinement rounds. The right panel reports the mean
    assignment-side uncertainty $\overline{s}_e^{(r)}$ and CATE-side
    uncertainty $\overline{s}_\tau^{(r)}$. Error bars and shaded
    regions indicate variation across runs.}
    \label{fig:refinement_dynamics_app}
\end{figure*}

Figure~\ref{fig:refinement_dynamics_app} shows that the cache expands
rapidly during the early rounds but exhibits progressively smaller
marginal growth. The final cumulative cache ratio ranges from $68.7\%$ to
$82.5\%$ on IHDP and from $51.2\%$ to $63.2\%$ on Jobs. Across all
eight dataset--host combinations, the cache ratio is unchanged
between the fourth and fifth refinement rounds, indicating that few
or no previously unqueried units are surfaced at that stage.

Meanwhile, both $\overline{s}_e^{(r)}$ and
$\overline{s}_\tau^{(r)}$ exhibit an overall downward trend, with
only minor intermediate fluctuations. The simultaneous reduction in
predictive uncertainty and saturation of the cache is consistent with
the estimator becoming locally more stable while the allocation set
approaches a fixed configuration. This dynamic evidence complements
the performance-based sensitivity analysis of $R$ in
Section~\ref{subsec:sensitivity_analysis}.

\section{Analysis of Semantic Representation}
\label{app:representation_lossiness}

\begin{figure}[t]
    \centering
    \includegraphics[width=\columnwidth]{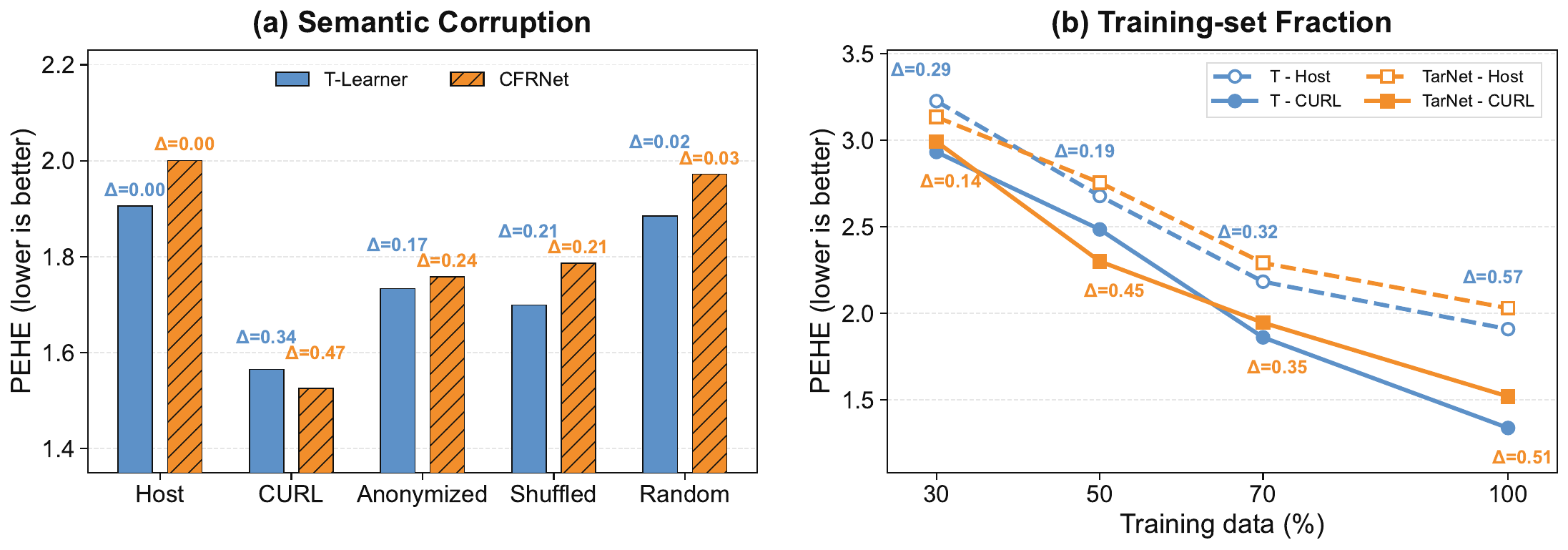}
    \caption{Additional validation of representational lossiness on IHDP. (a) Semantic-corruption results averaged over ten IHDP realizations and five seeds. Here,
    $\Delta=\mathrm{PEHE}_{\mathrm{Host}} -\mathrm{PEHE}_{\mathrm{Setting}}$, so positive values indicate improvement over the host. (b) PEHE under nested training-set fractions, averaged over ten realizations and three seeds. CURL improves both hosts at every fraction.}
    \label{fig:representation_lossiness}
\end{figure}

We conduct two complementary studies on IHDP to examine whether CURL's gains arise from pretrained representational structure and whether they persist across training-set sizes. Figure~\ref{fig:representation_lossiness} summarizes the results.

\paragraph{Semantic corruption.}
This experiment distinguishes gains from meaningful pretrained representations from those caused merely by additional vectors, parameters, or textual inputs. We use the ten IHDP realizations with T-Learner and CFRNet. Each realization is evaluated using seeds $\{42,52,62,72,82\}$, giving $10\times5=50$ results per bar.

The five settings are:
\begin{itemize}
    \item \textbf{Host}: the original host learner without CURL;
    \item \textbf{CURL}: the complete method with the original field and category semantics;
    \item \textbf{Anonymized}: field and category names are replaced by fixed identifiers such as \texttt{Feature F1} and \texttt{Category C1}, while values and category identities remain unchanged;
    \item \textbf{Shuffled}: category names are mapped to other names through a fixed within-field permutation while retaining the natural-language input format;
    \item \textbf{Random}: semantic embeddings are replaced by fixed random vectors of the same dimension, while the adapter, gates, and routing remain unchanged.
\end{itemize}

All settings use identical train/test splits, model seeds, and queried units. The uncorrupted CURL run first records its query plan, which is then replayed for the other semantic variants. We report mean PEHE and define
\begin{equation}
\Delta
=
\mathrm{PEHE}_{\mathrm{Host}}
-
\mathrm{PEHE}_{\mathrm{Setting}}.
\end{equation}
Thus, $\Delta>0$ indicates improvement over the host.

CURL achieves the lowest PEHE for both learners. Anonymized and Shuffled both degrade performance relative to CURL, while Random produces the largest deterioration. This ordered degradation indicates that CURL's improvement derives in part from the semantic structure encoded by the LLM representations, rather than merely from additional vectors, adapter parameters, or routing components.

\paragraph{Training-set fraction.}
We next examine whether CURL's advantage persists as the amount of training data changes. For each IHDP realization, we construct a fixed treatment-stratified split containing an 80\% training pool and a 20\% test set. From the same training pool, we form deterministic nested subsets at 30\%, 50\%, 70\%, and 100\%, such that each smaller subset is contained in the next larger subset.

The displayed comparison uses T-Learner and TARNet with model seeds $\{42,52,62\}$. Host and CURL use identical subsets, test sets, and model seeds. The LLM configuration is the same as in the semantic-corruption experiment. CURL obtains lower PEHE than its paired host for both learners at every training fraction.
Thus, CURL's benefit is not restricted to a particular subsample size and remains visible even when all available IHDP training data are used.

\section{Case Study: Semantic Interpretation of the Two Channels}
\label{app:case_study}

\begin{figure*}[t]
    \centering
    \includegraphics[width=0.8\textwidth]{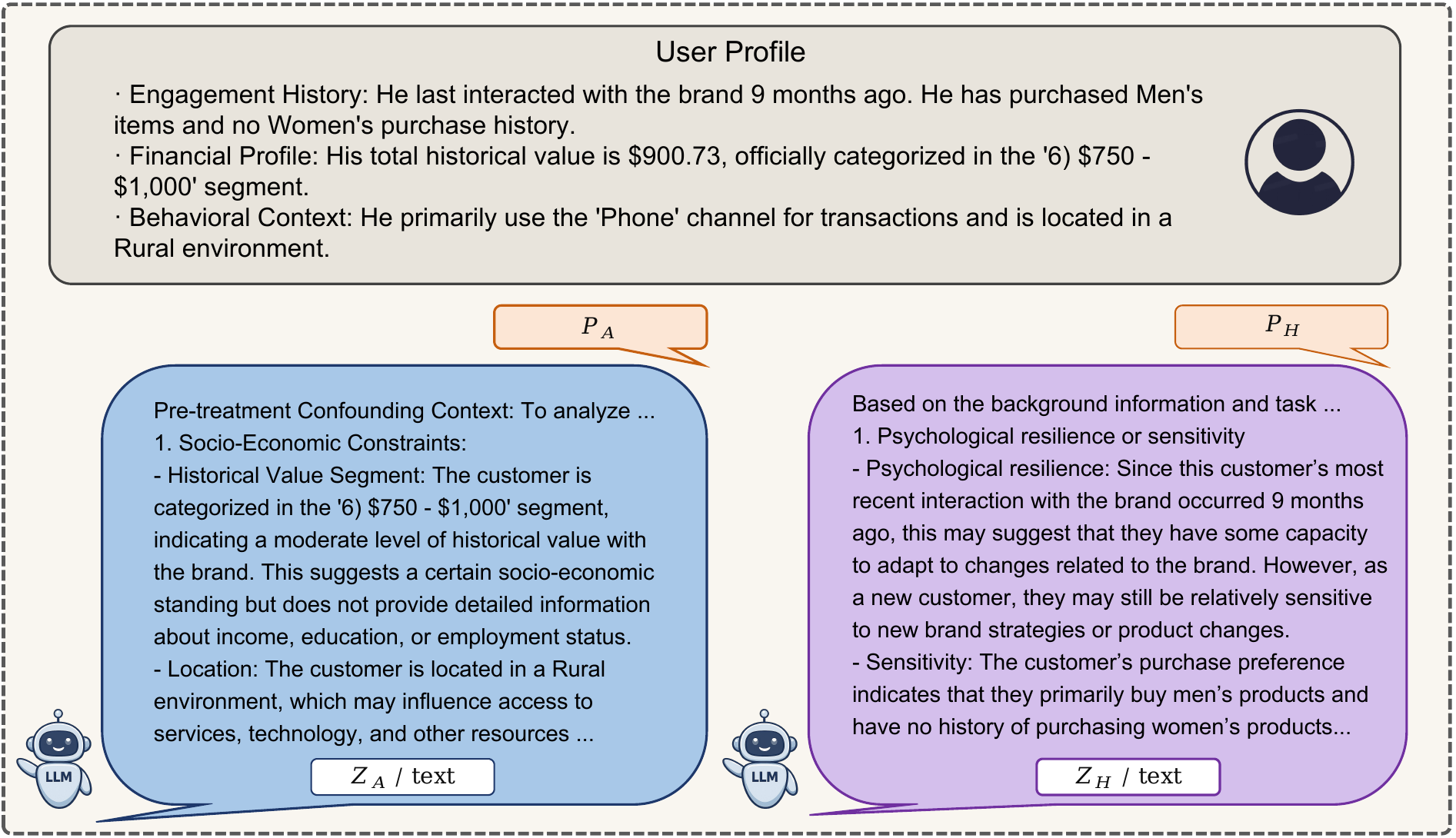}
    \caption{Walkthrough of a representative locally unstable
    Hillstrom unit. The observed profile, aggregate covariate-shift
    hint, assignment-oriented and heterogeneity-oriented textual
    traces, and downstream estimates illustrate how \textsc{CURL}
    reorganizes information already contained in the structured
    covariates.}
    \label{fig:case_study}
\end{figure*}

Figure~\ref{fig:case_study} examines a representative Hillstrom
customer selected by the uncertainty diagnostic. The observed profile
describes a relatively new customer who last interacted with the brand
nine months earlier, has \$900.73 in historical spending concentrated
on men's items, primarily transacts by phone, and resides in a rural
area. These fields are observed in $X$, but their semantic relations
are not explicit in the original numerical and categorical encoding.

The assignment-oriented trace emphasizes pre-treatment baseline
structure already supported by the profile, including the relation
between customer tenure, inactivity, accumulated value, transaction
channel, and geographic context. This trace should not be interpreted
as recovering hidden confounders or as implying non-random treatment
assignment in Hillstrom. Rather, it provides a semantic organization
of the observed baseline profile for the shared and assignment-side
components of the estimator.

The heterogeneity-oriented trace instead emphasizes relations that may
be useful for modeling differential response to the Men's E-Mail
campaign. In particular, the combination of men's-product purchase
history, prolonged inactivity, historical value, and preferred
transaction channel provides a more explicit description of potential
content match and engagement sensitivity. These are semantic
interpretations of observed fields rather than newly observed
individual characteristics.

The textual traces are displayed only for interpretation. The actual
adapter inputs are $z_{A,i}^{\mathrm{LLM}}$ and
$z_{H,i}^{\mathrm{LLM}}$, extracted from the corresponding generated
responses. The case study therefore illustrates how pretrained
semantic structure can reshape the effective representation of $X$
without changing the observed information set, treatment assignment,
outcome, estimand, or identifying assumptions.

\end{document}